%% file: main.tex
\documentclass[12pt]{article}
\usepackage[margin=1in,a4paper]{geometry}
\usepackage{parskip}
\usepackage{amsmath}
\usepackage{amssymb}
\usepackage{csquotes}
\usepackage{graphicx}
\usepackage{xcolor}
\usepackage{subcaption}
\usepackage{caption}
\usepackage{float}
\usepackage{tikz}
\usetikzlibrary{arrows.meta, positioning, shapes, decorations.pathreplacing, fit, backgrounds, calc}
\usepackage{booktabs}
\usepackage{array}
\usepackage{longtable}
\usepackage{multirow}
\usepackage{tabularx}
\usepackage{dcolumn}
\usepackage{etoolbox}
\AtBeginEnvironment{table}{\singlespacing}
\AtBeginEnvironment{longtable}{\singlespacing}

\usepackage{listings}
\usepackage{minted}
\usepackage{lscape}
\usepackage{soul}
\sethlcolor{lightgray}
\usepackage{authblk}
\usepackage[authordate, backend=biber, natbib, dashed=false]{biblatex-chicago}
\DeclareSourcemap{
  \maps[datatype=bibtex]{
    \map{
      \pertype{software}
      \step[typesource=software, typetarget=misc]
    }
  }
}
\AtEveryBibitem{%
  \ifboolexpr{
    test {\ifentrytype{article}} or
    test {\ifentrytype{inproceedings}} or
    test {\ifentrytype{incollection}} or
    test {\ifentrytype{book}} or
    test {\ifentrytype{techreport}} or
    test {\ifentrytype{unpublished}}
  }{%
    \iffieldundef{doi}{}{%
      \clearfield{url}%
      \clearfield{urldate}%
    }%
  }{}%
}
\usepackage[hidelinks]{hyperref}
\usepackage{xurl}
\usepackage{setspace}
\newif\ifanonymoussubmission
\anonymoussubmissionfalse
\ifdefined\ANONYMIZED
\anonymoussubmissiontrue
\fi
\ifdefstring{\jobname}{main-anonymized}{\anonymoussubmissiontrue}{}
\newcommand{\submissiondateblock}{%
\ifanonymoussubmission
\vfill
\else
\today \vfill
\fi
}
\newcommand{\submissionhref}[1]{%
\ifanonymoussubmission
\texttt{[link redacted for anonymous review]}%
\else
\href{__#1__}{\nolinkurl{#1}}%
\fi
}
\title{\vfill Magic Words or Methodical Work? Challenging Conventional Wisdom in LLM-Based Political Text Annotation}

\ifanonymoussubmission
\author{}
\else
\author[1]{Lorcan McLaren}
\author[1]{James P. Cross}
\author[1]{Zuzanna Krakowska}
\author[1]{Robin Rauner}
\author[2]{Martijn Schoonvelde}

\affil[1]{University College Dublin}
\affil[2]{University of Groningen}
\fi
\date{\submissiondateblock}

\begin{document}
\maketitle
\newpage
\vspace*{\fill}
\begin{abstract}
    Political scientists are rapidly adopting large language models (LLMs) for text annotation, yet the sensitivity of annotation results to implementation choices remains poorly understood. Most evaluations test a single model or configuration; how model choice, model size, learning approach, and prompt style interact, and whether popular ``best practices'' survive controlled comparison, are largely unexplored. We present a controlled evaluation of these pipeline choices, testing six open-weight models across four political science annotation tasks under identical quantisation, hardware, and prompt-template conditions. Our central finding is methodological: interaction effects dominate main effects, so seemingly reasonable pipeline choices can become consequential researcher degrees of freedom. No single model, prompt style, or learning approach is uniformly superior, and the best-performing model varies across tasks. Two corollaries follow. First, model size is an unreliable guide both to cost and to performance: cross-family efficiency differences are so large that some larger models are less resource-intensive than much smaller alternatives, while within model families mid-range variants often match or exceed larger counterparts. Second, widely recommended prompt engineering techniques yield inconsistent and sometimes negative effects on annotation performance. We use these benchmark results to develop a validation-first framework---with a principled ordering of pipeline decisions, guidance on prompt freezing and held-out evaluation, reporting standards, and open-source tools---to help researchers navigate this decision space transparently.
\end{abstract}
\vspace*{\fill}
\newpage
\pagenumbering{arabic}

\section{Introduction}

Large language models (LLMs) have been adopted as text annotation tools in political science faster than the evaluative tradition needed to distinguish principled guidance from annotation folklore. When researchers embraced dictionary methods and topic models, a substantial body of work evaluated how configuration choices including stopword lists, the number of topics, and seed words to name but a few, affected substantive conclusions \citep{grimmer_text_2013,grimmer_text_2022}. That evaluative tradition produced widely cited principles: that all models are wrong but some are useful, that no single method dominates, and that validation against human judgement is essential \citep{grimmer_text_2013}. No equivalent guidance yet exists for LLM-based annotation, even as these models are deployed to measure widely varying constructs, including policy positions \citep{mens_positioning_2025}, protest events \citep{halterman_codebook_2025}, and political rhetoric \citep{tornberg_large_2024}.

The absence of systematic evaluation criteria matters because a researcher's decision space is vast. Deploying an LLM as an annotator requires choosing a model family, a model size, a prompting strategy, and identifying whether or not to provide labelled examples. Each of these decisions interact with one another and with the properties of the annotation task itself. If these choices substantially affect results, then LLM-based annotation introduces a new form of \textit{researcher degrees of freedom}: undisclosed analytical flexibility that can shift findings toward preferred conclusions, whether intentionally or not \citep{baumann_large_2025}. Recent work quantifying ``LLM hacking'' demonstrates that minor prompt paraphrases alone can render virtually any hypothesis statistically significant \citep{baumann_large_2025}. Yet most applied studies select pipeline configurations on the basis of convention, informal ``best practice'' guides \citep{tornberg_best_2024}, or resource availability, without reporting how alternative choices would have affected their measurements.

This paper contributes to a growing body of work evaluating LLM annotation pipeline choices \citep{alizadeh_open-source_2024,atreja_whats_2025,halterman_codebook_2025,baumann_large_2025} by using a large controlled comparison to address a methodological problem in political text analysis: how researchers should validate LLM annotation pipelines when multiple defensible implementation choices are available. We systematically vary four choices that an analyst must make when employing an LLM in an annotation workflow: model choice (six open-weight models), model size (three model families with 270M--72B parameter variants), zero-shot versus few-shot learning, and prompt style (standard, persona, chain-of-thought). We explore the effect of these choices on annotation performance across four annotation tasks spanning different text types, annotation formats, and levels of conceptual complexity. Two clarifications about scope are important. We do not argue that generative LLMs are the superior approach to these tasks: fine-tuned encoder-only models such as BERT still outperform generative models on many text annotation tasks in absolute terms, including some of ours. Nor do we seek to identify optimal settings for each model--task combination. Rather, we use benchmark variation across configurations to show the \textit{relative} difference that implementation choices make, the at times unpredictable direction of these effects, and the implications this has for methodological practice.

Our focus on open-weight models is deliberate and inspired by both open-science and environmental concerns. Closed-source systems, while sometimes more performant, cannot guarantee reproducibility across time, preclude measurement of energy consumption, and may be modified or withdrawn without notice. This makes reproducibility hostage to the whims of private companies. By running all models locally, we can precisely quantify the computational cost of each design choice alongside its effect on annotation performance. Efficiency constraints, and energy use in particular, deserve legitimate attention: the inference costs of generative LLM pipelines can be orders of magnitude larger than those of earlier transformer models or other quantitative text analysis methods, and pipeline choices that appear inconsequential for performance can produce order-of-magnitude differences in energy consumption. All experiments use identical quantisation (Q4\_K\_M), identical hardware (NVIDIA A100 GPUs), and standardised prompt templates, isolating the marginal effect of each design choice. The full experimental pipeline, from codebook definition to LLM benchmarking, is released as open-source software to facilitate replication and extension (see Data and Code Availability).

Our experiments are designed to adjudicate a set of concrete heuristics that already circulate in the field: that larger models produce better annotations, that larger models are necessarily more resource-intensive, that providing labelled examples (few-shot) is uniformly beneficial, and that common forms of prompt engineering reliably improve performance. Three headline findings emerge. First, \textit{interaction effects dominate main effects}: no model, prompt technique, or learning approach is uniformly superior, and task-level variation routinely exceeds the average treatment effect of any single pipeline choice. Second, \textit{model size is an unreliable guide both to performance and to cost}: cross-family efficiency differences are so large that Gemma~3 at 27B parameters is less resource-intensive than the smallest tested variants of other families, while within model families annotation performance scales non-monotonically and mid-range variants sometimes outperform the largest available model. Third, \textit{popular prompt engineering techniques are unreliable}: both persona and chain-of-thought prompting produce erratic performance effects, sometimes improving and sometimes degrading performance, while potentially substantially increasing computational cost. Taken together, these results challenge not just the assumption that ``bigger is better,'' but also the assumption that bigger is necessarily more expensive to run.

These findings demonstrate that LLM annotation pipeline choices are a consequential form of researcher degrees of freedom, comparable in importance to those documented for earlier text-as-data methods. We therefore make two linked contributions. First, we provide the empirical evidence base that has been lacking: a controlled comparison showing how and why pipeline choices matter, and which existing ``best practice'' recommendations survive scrutiny. Second, we translate those results into a practical \textit{validation-first framework} (Section~\ref{sec:validation-framework}) that specifies how researchers should move from codebook design to model evaluation: freeze the annotation instructions, benchmark alternative pipelines against human-coded samples, and treat prompt revision as a tuning decision that requires transparent reporting and, where iterative development occurs, a separate held-out test set. Echoing the principles articulated by \citet{grimmer_text_2013} for earlier text-as-data methods, we argue that any LLM annotation pipeline should be pilot-tested against human-coded samples before deployment at scale.

\section{Related Literature}

\subsection{LLMs in the Text-as-Data Toolkit}
Large language models have demonstrated considerable promise for a variety of tasks in computational social science \citep[e.g.,][]{ziems_can_2023,bail_can_2024,thapa_large_2025,tornberg_chatgpt-4_2023,tornberg_large_2024}. Their versatility extends across multiple research applications, from text annotation to agent-based modeling to synthetic data generation.

These recent efforts are part of a longer tradition of text-as-data studies, where not just measurement but also evaluation techniques play a central role in the scientific process. \citet{grimmer_text_2013} articulated four foundational principles: all quantitative models of language are wrong but some are useful; no single method dominates; the validation of results against human judgement is essential; and quantitative text analysis augments rather than replaces careful reading. These principles were developed for dictionary, scaling, and topic-model methods, but they apply with equal, and arguably greater force to LLM-based annotation, where the parameter space of implementation choices is orders of magnitude larger.

Recent work has begun to extend this evaluative tradition to generative models. \citet{halterman_codebook_2025} propose a five-stage framework for ``codebook LLM'' measurement, demonstrating that current open-weight models (7--12B parameters) have significant limitations in following codebook instructions zero-shot but that supervised fine-tuning can substantially improve performance. \citet{mens_positioning_2025} show that instruction tuned LLMs can position political texts in ideological spaces with correlations exceeding 0.90 against expert benchmarks, while cautioning that empirical validation remains essential. \citet{argyle_out_2023} demonstrate that LLMs can simulate diverse human subpopulations in surveys, raising both opportunities and concerns about construct validity. In each case, the sensitivity of results to model selection and prompting choices is acknowledged but not systematically evaluated.

Most directly related to our study, \citet{alizadeh_open-source_2024} provide a practical guide to open-weight LLMs for text annotation, systematically varying model choice, temperature, zero-shot versus few-shot prompting, fine-tuning sample size, and model scale across political science tasks. Our study builds on this foundation by extending to more recent model families (released through mid-2025), adding prompt style variation (persona and chain-of-thought), measuring computational efficiency alongside performance, and holding all models to identical quantisation and hardware conditions to isolate pipeline effects from infrastructure confounds.

Our contribution differs from these prior studies in that we evaluate the \textit{pipeline choices themselves} as objects of inquiry under tightly controlled conditions. Rather than demonstrating that a particular LLM can perform a particular task, we ask how much the choice of model, size, prompting strategy, and learning approach matters, and whether the popular heuristics researchers use to navigate these choices are empirically justified.

\subsection{Model Choice}

Most existing work on LLM-based annotation in political science has used closed-source models such as GPT-3.5 and GPT-4 from OpenAI \citep[e.g.,][]{tornberg_large_2024,heseltine_large_2024}. While these models offer advantages, lower barriers to entry, reduced technical expertise requirements, minimal computational needs, high-quality documentation, and strong performance—their use raise significant concerns for reproducibility and transparency. Research exploring open-weight alternatives has typically focused on older or smaller models \citep[e.g.,][]{halterman_codebook_2025}, creating unfair comparisons with state-of-the-art closed-source systems \citep[e.g.,][]{ziems_can_2023,atreja_prompt_2024,barrie_prompt_2025,barrie_replication_2024}.

While \citet{halterman_codebook_2025} evaluate open-weight models in the 7--12B parameter range and \citet{alizadeh_open-source_2024} test models up to Llama-2 70B, our study extends the comparison to models released through mid-2025 (including Qwen~3, Gemma~3, and GPT~OSS), providing a more current picture of the open-weight landscape. Crucially, we hold quantisation level, hardware, and prompt structure constant across all comparisons, isolating model choice from confounding implementation differences that complicate cross-study comparison.

\subsection{Model Size}

Language model size (measured in parameters) significantly impacts performance, particularly for in-context learning. In-context learning refers to the ability of large language models (LLMs) to learn and perform new tasks from examples or instructions provided within the prompt itself, without updating the model's parameters \citep{dong_survey_2024}. Larger models demonstrate superior ability to override semantic priors and learn contradictory input-label mappings, while smaller models rely more heavily on priors \citep{wei_larger_2023}. Nevertheless, even small models can effectively learn from examples \citep{schick_its_2021}. 

Systematic investigation of the effect of model size on annotation performance is lacking in the literature, as researchers typically opt instead to compare different model choices. However, model size has significant implications for the energy use involved in inference \citep{samsi_words_2023}.  Furthermore, much existing work involving open-weight models employs small variants (7-8B parameters), which do not provide a reasonable comparison to much larger closed-source models accessible via API (typically in the range of \textgreater100B parameters).

The assumption that larger models perform better draws implicitly on neural scaling laws, which suggest predictable relationships between model size, training compute, and pre-training loss \citep{kaplan_scaling_2020,hoffmann_training_2022}. However, pre-training loss is a poor proxy for downstream annotation performance on domain-specific tasks \citep{halterman_codebook_2025}. Instruction tuning, task-model compatibility and model shrinking techniques such as pruning and distillation may introduce non-linearities that violate the smooth scaling curves observed in pre-training. Our within-family size comparisons provide a direct test of whether scaling intuitions hold for political science annotation tasks.

\subsection{Zero-Shot vs. Few-Shot Learning}
Zero-shot and few-shot learning represent the two dominant paradigms for deploying LLMs as text annotators without task-specific fine-tuning. In zero-shot settings, models rely solely on instructions and their pre-trained knowledge, while few-shot approaches additionally provide labelled examples within the prompt to guide annotation \citep{brown_language_2020}. Few-shot methods consistently outperform their zero-shot counterparts across political science applications. For instance, \citet{laurer_less_2024} demonstrate that BERT-NLI models trained on as few as 500 examples can match the performance of classical supervised models trained on ten times as much data, while \citet{burnham_political_2024} show that domain-adapted models achieve state-of-the-art few-shot annotation of political texts with orders of magnitude greater computational efficiency than large commercial LLMs. However, performance gains from example provision vary systematically by task complexity. Binary and low-cardinality annotation tasks with semantically distinct categories tend to benefit most, whereas fine-grained taxonomies with many categories show more modest improvements \citep{halterman_few-shot_2021,wen_zero-shot_2023}. Recent work also suggests that in-context learning consistently outperforms instruction tuning across computational social science tasks \citep{wang_instruction_2024}, and that the quality and representativeness of provided examples matters more than their quantity \citep{wang_instruction_2024,timoneda_memory_2025}.

Despite these advances, most studies examining zero-shot and few-shot performance do so incidentally, evaluating one approach or the other, rather than systematically comparing the two under controlled conditions across multiple tasks. Furthermore, the computational cost of few-shot prompting is rarely quantified: longer prompts increase inference time and energy consumption, yet these trade-offs are seldom reported alongside performance metrics.

\subsection{Prompt Style}
LLMs have shown sensitivity to slight variations in prompt formatting \citep{sclar_quantifying_2024}, with minor changes potentially causing substantial impacts on performance. However, larger models generally demonstrate greater robustness to these variations \citep{he_does_2024}. Two widely adopted approaches to LLM prompting that are thought to improve model outputs are (1) persona prompting, and (2) chain-of-thought prompting. 

\subsubsection{Persona Prompting}
Persona prompting where the analyst provides an LLM with a character description to role play, represents a common technique in the field. This approach has been used to measure demographic stereotypes and bias embedded in LLMs \citep{cheng_marked_2023} and to simulate diverse samples of survey respondents \citep{argyle_out_2023}. Despite its widespread adoption among social scientists, evidence for its effectiveness in annotation tasks remains less compelling.

While persona prompting may occasionally lead to performance gains, the effect of each persona can be largely random \citep{zheng_when_2024}. Other studies suggest that persona prompting offers modest yet significant improvements across some tasks, with the most pronounced benefits occurring in cases where annotators largely disagree but only by a small margin \citep{hu_quantifying_2024}. These findings suggest that the benefits of persona prompting for text annotation may be limited and task-dependent, motivating systematic empirical investigation of its effects under controlled conditions.

\subsubsection{Chain-of-Thought Prompting}
Chain-of-thought (CoT) prompting \citep{wei_chain--thought_2022} encourages models to articulate their reasoning process before providing final answers. While this approach is not necessarily a reliable path to interpretability \citep{turpin_language_2023}, it does appear to induce some form of reasoning. Despite this, CoT prompting does not always improve performance \citep{savelka_can_2023}, particularly for text annotation tasks where clear decision boundaries are more important than complex reasoning chains.

CoT implementations vary considerably, with some researchers simply encouraging models to "think step-by-step" \citep[e.g.,][]{kojima_large_2022}, while others require explicit explanations of reasoning either before or after annotation decisions, and provide reasoning steps alongside their example sets \citep{wei_chain--thought_2022}. Recent research demonstrates that task-specific prompting significantly outperforms unsupervised prompt generation, emphasising the necessity of thoughtful human guidance in CoT prompting \citep{zhang_why_2025}. However, CoT prompting does not improve performance on language reasoning to the same extent as it does with arithmetic tasks \citep{wang_instruction_2024}. Despite its theoretical appeal, the computational overheads of CoT prompting raise important questions about efficiency trade-offs when undertaking large-scale text annotation tasks.

Most closely related to our work, \citet{atreja_whats_2025} conduct a large-scale experiment testing how model selection and prompt design features (definition inclusion, output type, explanation requests, and prompt length) affect LLM annotation compliance and accuracy across four computational social science tasks. They find that compliance and accuracy are prompt-dependent and that minor changes can cause large shifts in label distributions. Our study extends this line of inquiry in two directions: we cross prompt style with model choice (testing all style--model combinations rather than evaluating each in isolation) and we systematically measure the efficiency costs of prompt engineering techniques, providing a performance-per-unit-cost perspective absent from prior work.

\section{Methods}
This paper employs a controlled factorial design to measure the performance of LLMs on four distinct text annotation tasks from political science. By systematically varying four dimensions of the annotation pipeline (model choice, model size, learning approach, and prompt style), while holding hardware, quantisation, and prompt structure constant, the design isolates the marginal contribution of each factor and reveals interaction effects that single-dimension evaluations cannot detect. One of the four tasks is an original project involving one of the co-authors, while the remaining three are replications of existing papers. The tasks are deliberately chosen to span a wide range of text types (parliamentary speeches, newspaper excerpts, transcriptions of committee deliberations), response formats (binary labels, multi-class categorisation, ordinal Likert scales), and units of analysis (sentences, article segments, debate interventions, full speeches). The conceptual complexity of the tasks also varies considerably: from economic sentiment, which requires a relatively straightforward assessment of tone, to manifesto topic annotation, which relies on a domain-specific coding scheme with definitions and examples that must be absorbed through in-context learning. Together, these tasks provide a robust testbed for evaluating LLM annotation performance under diverse conditions.

\subsection{Task Definitions}\label{sec:task-definitions}
Each of the tasks is briefly introduced below, and Table~\ref{tab:task-overview} provides a summary overview. See the codebooks in Appendix~\ref{app:codebooks} for a complete overview of the tasks, definitions for each dimension, and example sets. For the psychological distance task, the codebook reflects the exact coding instructions distributed to human annotators. For the three replication tasks, codebooks were reconstructed from the appendices and supplementary materials of the original papers, including the examples used to illustrate category boundaries where these were available. For the psychological distance task, the examples are the same ones used in the human coding process: they were developed through iterative labelling rounds and intercoder-reliability checks to clarify recurrent disagreements, then frozen once the codebook was finalised. Codebooks for all four tasks were formalised using CodeBook Studio \citep{mclaren_codebook_studio_2026}, a web-based annotation application that supports binary, categorical, Likert, and free-text annotation types and provides structured outputs. For the psychological distance task, CodeBook Studio was also used to collect the human annotations; for the remaining three tasks, ground-truth data were obtained from the original authors' published materials.

\subsubsection{Approval}
General approval \citep{wratil_public_2019} refers to the degree to which a speaker expresses approval of the proposal(s) being negotiated or the state of negotiations. The data consists of transcriptions of video footage of public deliberations of the Council of the EU's ECOFIN configuration (2010--2015), with the debate participation as the unit of analysis. Coders first identify the major dimension of contestation within each debate, typically either the legislative proposal presented by the European Commission, the Council's state of play on the proposal, or the state of negotiations with the European Parliament, and then assess each speaker's degree of approval on a 5-point Likert scale, where 1 indicates full approval and 5 indicates full disapproval. The ground truth comprises 1,110 debate participations by national delegations, the Commission, and other actors (e.g.\ the ECB) across 86 debates, coded by a single coder; a random 25\% of the dataset was double-coded to ensure sufficient intercoder reliability ($\alpha = 0.82$). The data is right-skewed: an approval score of 1 accounts for 37\% of observations, with the proportion decreasing gradually as the scale increases, and a score of 5 accounting for just 5\%. Prior research shows that asking LLMs to provide numerical scores, as in the case of a Likert scale, instead of labels reduces compliance and accuracy \citep{atreja_whats_2025}.

\subsubsection{Psychological Distance}
The psychological distance of climate change refers to the \textit{sense of proximity or distance} from the impacts of climate change created by a speech or text \citep{mclaren_here_2025}. The data consists of European Parliament plenary speeches (2014--2024), with the sentence as the unit of analysis. The task is assessed in a three-step process: (1) determining whether references to environmental impacts are present [binary]; (2) assessing their specificity [categorical: specific; universal]; and (3) evaluating their implied proximity [categorical: proximate; distant]. The ground truth comprises 1,037 sentences, coded by three human annotators using the codebook in Appendix~\ref{app:codebooks}; final annotations for each sentence were resolved through majority agreement. Because the three annotation dimensions are nested, each successive step applies to a smaller subset of the data. The dataset is heavily imbalanced at the first step: 89.2\% of sentences contain no reference to environmental impacts (Presence = No), leaving only 10.8\% for subsequent annotation. Of those, roughly 60\% are coded as Universal and 40\% as Specific. Among the Specific subset, approximately two-thirds are coded as Proximate and one-third as Distant.

\subsubsection{Economic Sentiment}
Economic sentiment refers to the tone of newspaper coverage of U.S.\ national economic performance \citep{barbera_automated_2021}. The data consists of newspaper articles, with the article segment (first five sentences) as the unit of analysis. Each segment is annotated as either negative or positive [binary] based on the indication the text gives about how the U.S.\ economy is performing. The ground truth comprises 420 segments, each coded by a minimum of seven CrowdFlower coders on a 9-point ordinal scale (1 = very negative, 9 = very positive). The average tone across coders was computed for each segment and then dichotomised: scores of 1--4 were coded as negative (0) and scores of 6--9 as positive (1), while segments with an average score of 5 were discarded. The resulting dataset is relatively balanced, with 39\% positive and 61\% negative instances.

\subsubsection{Manifesto Topic}
The manifesto topic task assigns political text to policy topic categories based on the Comparative Manifesto Project (CMP) coding scheme \citep{osnabrugge_cross-domain_2023}. The data consists of speeches from the New Zealand parliament (1987--2002), with the speech as the unit of analysis. Following \citet{osnabrugge_cross-domain_2023}, we use a collapsed 8-topic specification rather than the full set of 44 Manifesto Project topics, as each topic is provided with its own definition and examples drawn from the original Manifesto Project codebook and including all 44 would excessively lengthen the prompt. Each speech is assigned to one of eight domain categories [categorical: No Domain, External Relations, Freedom And Democracy, Political System, Economy, Welfare And Quality Of Life, Fabric Of Society, and Social Groups]. The ground truth comprises 4,165 speeches, coded by a single coder from New Zealand trained by the Manifesto Project team; the coder's performance was validated by comparing their annotations with those of three other trained coders on a random sample of 250 speeches. The dataset exhibits significant class imbalance: Political System and Welfare and Quality of Life account for 26\% and 19\% of all speeches respectively, while No Topic comprises 4.6\% and External Relations just 2.3\%.

\begin{table}[H]
    \centering
    \footnotesize
    \setlength{\tabcolsep}{4pt}
    \renewcommand{\arraystretch}{1.15}
    \begin{tabularx}{\linewidth}{@{}>{\raggedright\arraybackslash}p{1.55cm}>{\raggedright\arraybackslash}p{1.8cm}>{\raggedright\arraybackslash}p{1.7cm}>{\raggedright\arraybackslash}p{1.7cm}>{\centering\arraybackslash}p{0.8cm}>{\centering\arraybackslash}p{1.25cm}>{\raggedright\arraybackslash}X>{\raggedright\arraybackslash}p{2cm}@{}}
        \toprule
        \textbf{Task} & \textbf{\shortstack[l]{Text\\Type}} & \textbf{\shortstack[l]{Unit of\\Analysis}} & \textbf{\shortstack[l]{Annotation\\Type(s)}} & \textbf{$N$} & \textbf{\shortstack[l]{No.\\Coders}} & \textbf{\shortstack[l]{Aggregation\\Strategy}} & \textbf{Source} \\
        \midrule
        Approval & Council of EU deliberations & Debate participation & 5-pt Likert & 1,110 & 1 (25\% double-coded) & Single coder & \citet{wratil_public_2019} \\
        \addlinespace
        Psych. Distance & EP plenary speeches & Sentence & Binary, categorical (2 classes) & 1,037 & 3 & Majority agreement & Original \\
        \addlinespace
        Econ. Sentiment & Newspaper articles & Article segment (first 5 sent.) & Binary & 420 & $\geq$ 7 & Averaged 9-pt scale, dichotomised & \citet{barbera_automated_2021} \\
        \addlinespace
        Manifesto Topic & NZ parl.\ speeches & Speech & Categorical (8 classes) & 4,165 & 1 (validated against 3) & Single coder & \citet{osnabrugge_cross-domain_2023} \\
        \bottomrule
    \end{tabularx}
    \caption{Overview of the four annotation tasks. $N$ = number of units in the ground truth dataset.}
    \label{tab:task-overview}
\end{table}

\subsection{LLM Annotation}

Our research explores the performance of various approaches to using language models as annotators with minimal training data, validated against human performance. We also investigate the trade-offs involved in using larger models and longer prompts in terms of inference time, energy intensity, and response length.

All models in our experiments are Q4\_K\_M quantised and run on 2 x NVIDIA A100-PCIE-40GB GPUs in the University College Dublin (UCD) Sonic HPC cluster. Quantisation techniques offer promising efficiency improvements, with 4-bit quantised models reducing energy consumption by approximately 3.5×, potentially cutting the overall energy footprint of inference by up to 72\% without sacrificing performance \citep{dettmers_qlora_2023}. We use each model's default sampling parameters as specified in its model card; where a model does not specify a parameter, Ollama's defaults of temperature = 0.8, top\_k = 40, and top\_p = 0.9 apply. Table~\ref{tab:model-comparison} reports the sampling parameters used for each model. Experiments are orchestrated by CodeBook Lab \citep{mclaren_codebook_lab_2026}, an open-source pipeline that takes the structured outputs of CodeBook Studio and benchmarks LLM performance across user-specified combinations of model, prompt style, and learning approach. The experimental grid is defined in a single configuration file, and inference is handled by Ollama. Energy consumption is tracked via CodeCarbon \citep{courty_mlco2codecarbon_2024}.

Each experimental dimension is tested while holding the remaining dimensions constant. The model choice and model size experiments use zero-shot prompting with the standard prompt style, isolating the effect of the model itself. The zero-shot versus few-shot comparison likewise uses the standard prompt style, varying only the provision of examples. The prompt style experiments include few-shot examples in all conditions, so that observed differences can be attributed to the prompt style rather than being confounded with the presence or absence of examples.

We structure prompts for different annotation types following standardised templates, as shown in Table \ref{tab:prompt-templates}. Prompt content (definitions, response options, and examples) is drawn directly from the codebooks in Appendix~\ref{app:codebooks}. These task-specific materials were frozen prior to benchmarking: no examples were added, removed, or rewritten on the basis of model outputs. The few-shot examples are therefore the same examples embedded in the corresponding human codebooks, and the persona and chain-of-thought conditions are implemented as pre-specified template variants layered on top of those frozen task instructions.

\begin{table}[H]
\centering
\begin{tabular}{|>{\raggedright\arraybackslash}p{3cm}|>{\raggedright\arraybackslash}p{12cm}|}
\hline
\textbf{Annotation Type} & \textbf{Standard Prompt Template} \\
\hline
Binary &
\begin{tabular}[t]{@{}p{12cm}@{}}
$<$section name$>$ \\
$<$section instructions$>$ \\
$<$annotation name$>$ \\
$<$tooltip$>$ \\
Respond with 1 if ``Yes" or 0 if ``No". \\
Return your response in JSON format, with the key ``response". \\
$<$example(s)$>$ \\
--- \\
Text: \\
``$<$text to annotate$>$" \\
Response:
\end{tabular} \\
\hline
Categorical &
\begin{tabular}[t]{@{}p{12cm}@{}}
$<$section name$>$ \\
$<$section instructions$>$ \\
$<$annotation name$>$ \\
$<$tooltip$>$ \\
Respond with $<$option1$>$, or $<$option2$>$, or ... \\
Return your response in JSON format, with the key ``response". \\
$<$example(s)$>$ \\
--- \\
Text: \\
``$<$text to annotate$>$" \\
Response:
\end{tabular} \\
\hline
Likert scale &
\begin{tabular}[t]{@{}p{12cm}@{}}
$<$section name$>$ \\
$<$section instructions$>$ \\
$<$annotation name$>$ \\
$<$tooltip$>$ \\
Respond with a whole number from $<$min value$>$ to $<$max value$>$ (inclusive), where $<$min value$>$ means lowest and $<$max value$>$ means highest. \\
Return your response in JSON format, with the key ``response". \\
$<$example(s)$>$ \\
--- \\
Text: \\
``$<$text to annotate$>$" \\
Response:
\end{tabular} \\
\hline
\end{tabular}
\caption{Standardised prompt templates by annotation type.}
\label{tab:prompt-templates}
\end{table}

Our experimental design explores the following dimensions:

\subsubsection{Model Choice}

We evaluate six open-weight language models as text annotators: Qwen 3 \citep{yang_qwen3_2025}, Qwen 2.5 \citep{qwen_qwen25_2025}, Gemma 3 \citep{team_gemma_2025}, GPT OSS \citep{openai_gpt-oss-120b_2025}, DeepSeek R1 \citep{deepseek-ai_deepseek-r1_2025}, and Llama 3.1 \citep{grattafiori_llama_2024}. Each model represents different architectural approaches and training methodologies. We assess their performance across all four annotation tasks to identify strengths, weaknesses, and efficiency trade-offs. Table~\ref{tab:model-comparison} summarises the key characteristics of each model.

\begin{table}[H]
    \centering
    \footnotesize
    \setlength{\tabcolsep}{4pt}
    \renewcommand{\arraystretch}{1.1}
    \begin{tabular*}{\linewidth}{@{\extracolsep{\fill}} l c l c c c c c @{}}
        \toprule
        \textbf{Model} & \textbf{Params.} & \textbf{Architecture} & \textbf{Release} & \textbf{Reason.} & \textbf{Temp.} & \textbf{Top-$k$} & \textbf{Top-$p$} \\
        \midrule
        Qwen 3 & 32B & Dense transformer & May 2025 & Yes & 0.6 & 20 & 0.95 \\
        Qwen 2.5 & 72B & Dense transformer & Sep.\ 2024 & No & 0.8* & 40* & 0.9* \\
        Gemma 3 & 27B & Dense transformer & Feb.\ 2025 & No & 1.0 & 64 & 0.95 \\
        GPT OSS & 20B (3.6B active) & MoE transformer & Aug.\ 2025 & Yes & 1.0 & 40* & 0.9* \\
        DeepSeek R1 & 70B & Dense transformer\textsuperscript{\textdagger} & Jan.\ 2025 & Yes & 0.8* & 40* & 0.9* \\
        Llama 3.1 & 70B & Dense transformer & Jul.\ 2024 & No & 0.8* & 40* & 0.9* \\
        \bottomrule
    \end{tabular*}
    \caption{Comparison of the six language models used in this study. All models are run in Q4\_K\_M quantisation. \textsuperscript{\textdagger}Distilled from DeepSeek R1 (671B MoE); the 8B and 70B variants use Llama base models, while smaller variants use Qwen 2.5 bases. * = Ollama default (parameter not specified in model card).}
    \label{tab:model-comparison}
\end{table}

\subsubsection{Model Size}

To understand the relationship between model size and annotation performance, we test model variants with different numbers of parameters from the Gemma 3, DeepSeek R1, and Qwen 3 families. This analysis tests whether larger models are in fact better at annotation tasks, and how the performance--efficiency tradeoff evolves as model size increases within a family.

\subsubsection{Zero-shot vs. Few-Shot Learning}

We compare zero-shot performance (using only instructions) against few-shot approaches (including instructions and examples) across our tasks. This comparison mirrors our manual annotation protocol, where human annotators received training examples. We measure whether and in what manner the provision of examples affects both performance and computational requirements.

\subsubsection{Prompt Style}

We examine three distinct prompt styles to determine their impact on annotation performance and model efficiency:

\begin{itemize}
    \item \textbf{Standard:} This baseline approach consists of three components: (1) a definition that establishes the concept to be measured, (2) a task description that specifies what the model should do with the input text, and (3) a set of response options that constrain the output format to ensure consistency.
    \item \textbf{Persona:} This approach adds a prefix to the standard prompt that describes specific attributes or expertise the model should embody while completing the task. We test whether framing the model with relevant expertise improves performance.
    \item \textbf{Chain-of-thought:} This approach appends a suffix to the standard prompt that encourages the model to make explicit use of its reasoning abilities before providing the final annotation. We investigate whether promoting reasoning improves performance, particularly for complex or ambiguous cases.
\end{itemize}

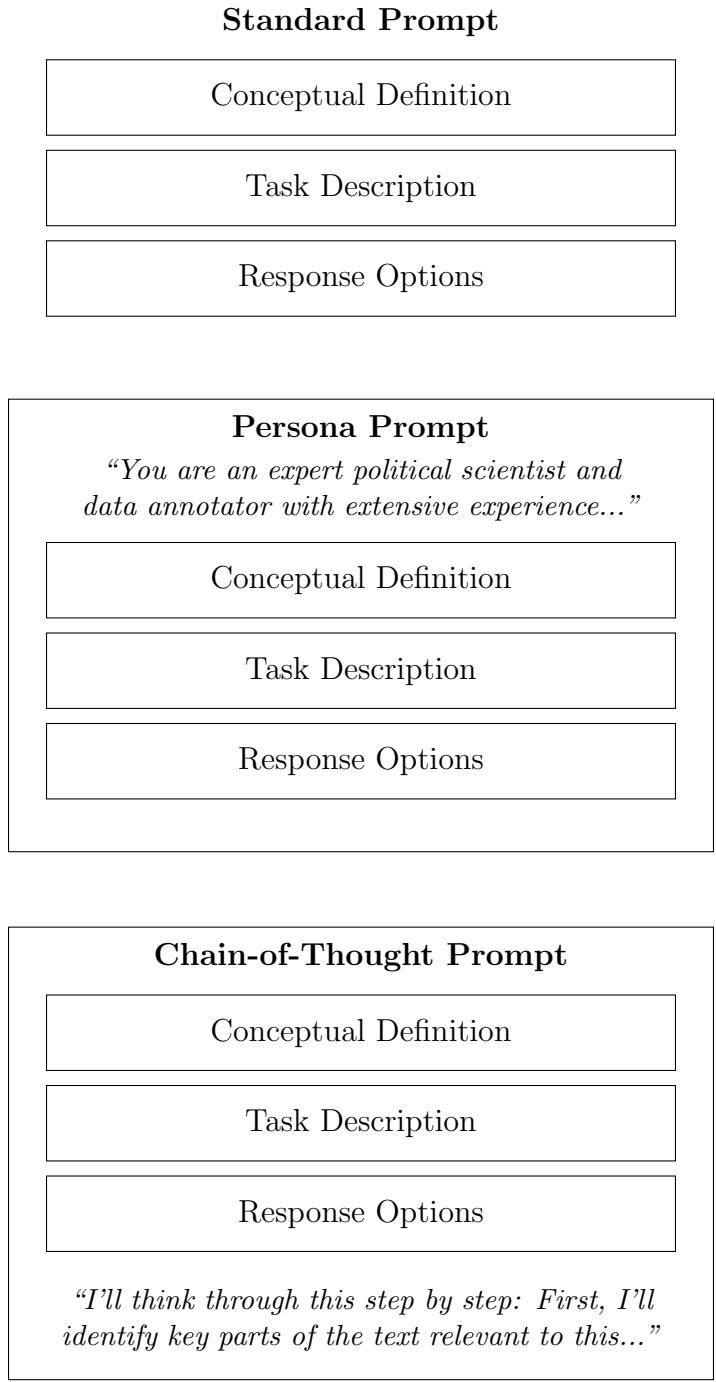
\begin{figure}[H]
\centering
\begin{tikzpicture}[
    block/.style={rectangle, draw, text width=8cm, text centered, minimum height=1cm},
    outerblock/.style={rectangle, draw, text width=9cm, minimum height=6cm}
    ]

\node[font=\bfseries] at (0,5.4) {Standard Prompt};
\node[block] (def) at (0,4.4) {Conceptual Definition};
\node[block] (task) at (0,3.2) {Task Description};
\node[block] (resp) at (0,2) {Response Options};

\node[font=\bfseries] at (0,0) {Persona Prompt};

\node[outerblock] (persona_outer) at (0,-2.6) {};

\node[font=\small\itshape, text width=8cm, align=center] (persona_example) at (0,-0.8) {``You are an expert political scientist and data annotator with extensive experience...''};

\node[block] (def_p) at (0,-2) {Conceptual Definition};
\node[block] (task_p) at (0,-3.2) {Task Description};
\node[block] (resp_p) at (0,-4.4) {Response Options};

\node[font=\bfseries] at (0,-7) {Chain-of-Thought Prompt};

\node[outerblock] (cot_outer) at (0,-9.6) {};

\node[block] (def_c) at (0,-8) {Conceptual Definition};
\node[block] (task_c) at (0,-9.2) {Task Description};
\node[block] (resp_c) at (0,-10.4) {Response Options};

\node[font=\small\itshape, text width=8cm, align=center] (cot_example) at (0,-11.8) {``I'll think through this step by step: First, I'll identify key parts of the text relevant to this...''};

\end{tikzpicture}
\caption{Structure of different prompt styles for LLM annotation}
\label{fig:prompt-styles}
\end{figure}

Figure \ref{fig:prompt-styles} illustrates the structure of each prompt style. We evaluate each approach to determine whether common prompt engineering techniques yield better performance on our annotation tasks, and at what computational cost.

\subsection{Evaluation}

Following the validation principles articulated by \citet{grimmer_text_2013}, we treat human annotations as the ground truth against which LLM performance is evaluated. Our evaluation approach varies by annotation type to ensure appropriate assessment of model performance. In all cases, the human-annotated sample described in Table~\ref{tab:task-overview} serves as validation data.

For binary or categorical annotations, we use classic classification metrics including accuracy, precision, recall, and F1 score. We also measure intercoder reliability between LLM annotations and ground truth through Cohen's kappa and Krippendorff's alpha, which allow direct comparison to human annotator agreement levels.

For Likert scale annotations, we employ both classic classification metrics and measures specifically designed for ordinal data. Quadratic weighted kappa and Spearman's correlation are particularly relevant for ordinal scales, as they penalise predictions that are further from the ground truth value more heavily than those that are closer.

Researchers face real computational and budgetary constraints when deploying LLMs at scale: annotation projects involving tens of thousands of documents may require days of GPU time and significant energy expenditure. To quantify these practical constraints, we measure three efficiency metrics for each model configuration, each capturing a distinct researcher concern:

\begin{itemize}
    \item \textbf{Energy consumption} (kilowatt-hours), estimated using CodeCarbon \citep{courty_mlco2codecarbon_2024}, captures the sustainability cost of inference.
    \item \textbf{Total output characters} captures the volume of text a model generates. Because API platforms charge per token, output length is a direct proxy for the financial cost of annotation at scale.
    \item \textbf{Inference time} (average per query and total) captures the practical workflow constraint, particularly relevant when annotating large corpora under deadline pressure.
\end{itemize}

We report these metrics separately rather than collapsing them into a composite index because our results show they are heterogeneous: a model can be fast but verbose (high token cost, low time cost), or energy-efficient but slow. Collapsing them would obscure the tradeoffs researchers need to evaluate against their own constraints. A researcher on a tight budget may prioritise minimising token volume; one running on local hardware may prioritise inference time; one concerned with environmental impact may prioritise energy. The normalised multi-panel figures we present let each reader weight these dimensions according to their own priorities.

All experiments were conducted on the UCD Sonic HPC cluster in Ireland, using NVIDIA A100-PCIE-40GB GPUs.\footnote{The Sonic cluster contains a heterogeneous set of GPU nodes (including NVIDIA V100, A100, and H100 graphics cards).} These measurements allow us to examine tradeoffs between annotation performance and computational cost, drawing conclusions about the relative merits of using larger models or more complex prompts under realistic resource constraints.

\section{Results}
The results are organised by experimental dimension. Within each subsection, we present an overall performance-to-efficiency comparison, decomposed efficiency metrics, and a task-level performance breakdown that reveals the interaction effects central to our argument. Together, these findings provide the empirical foundation for the validation-first framework we develop in Section~\ref{sec:validation-framework}. All metrics are macro-averaged, assigning equal weight to all classes for each response item. For tasks containing multiple response items of the same annotation type, we report the average metric across these items.

\subsection{Model Choice}

Applied studies often select a model on the basis of benchmark rankings or community reputation, implicitly assuming that a model that performs well on general-purpose tasks will also perform well as a text annotator. We test this by comparing six open-weight models of varying size and architecture on four annotation tasks.

\begin{figure}[H]
\centering
\includegraphics[width=\linewidth]{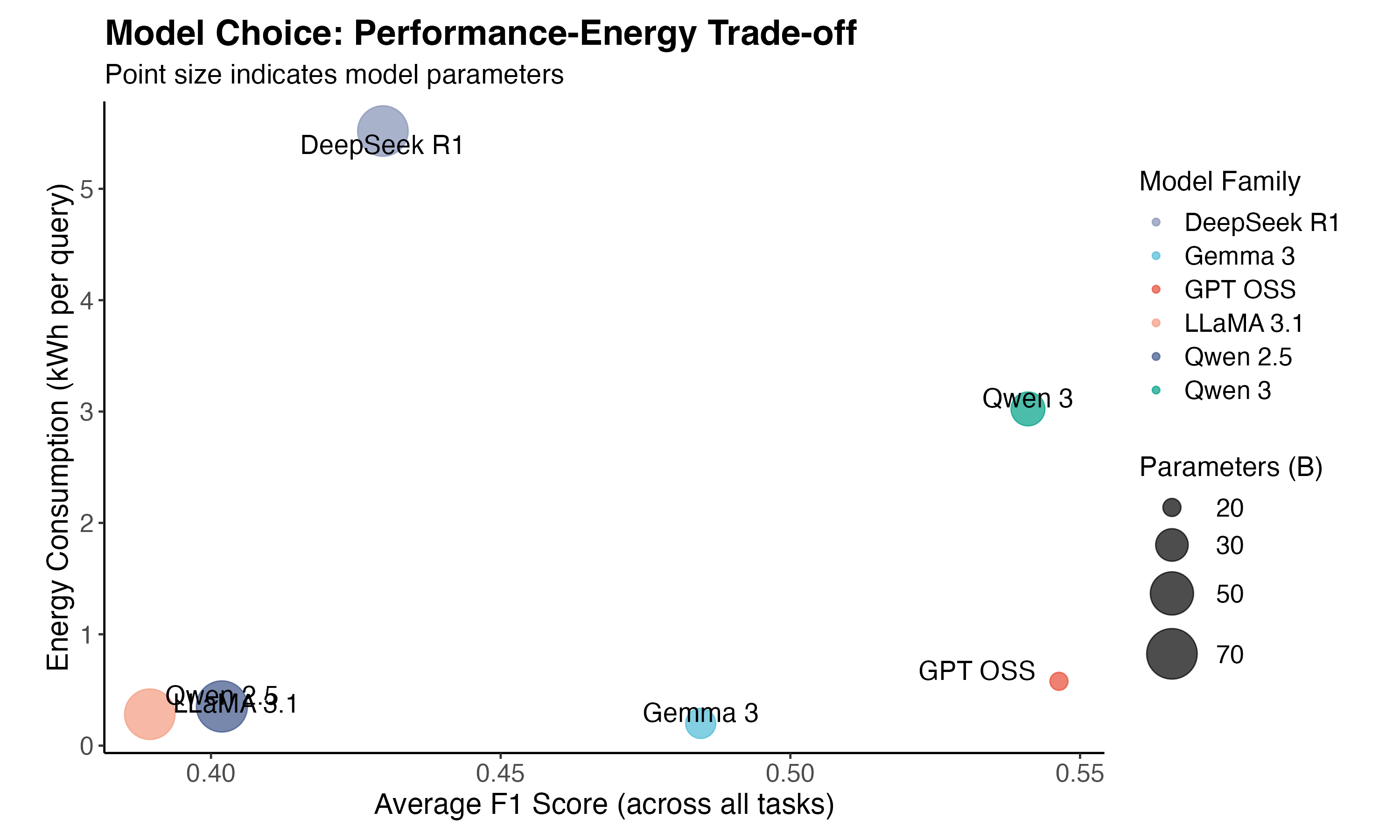}
\caption{Model choice performance--energy trade-off. Points show average F1 score and energy consumption for each model across the four annotation tasks, with point size indicating model parameters.}
\label{fig:model-choice-tradeoff}
\end{figure}

Figure \ref{fig:model-choice-tradeoff} plots the tradeoff between average annotation performance (F1) and energy consumption for each model. The ideal position is the lower-right corner: high performance at low energy cost. No model occupies this position convincingly, and no clear relationship between model efficiency, size and/or performance is observed. GPT OSS and Qwen 3 are effectively tied as the strongest performers overall, even though GPT OSS is the smallest model by total parameter count (20B). By contrast, the weakest performers are LLaMA 3.1 and Qwen 2.5, both of which underperform GPT OSS while using substantially more parameters. DeepSeek R1, although the same size as Llama 3.1, consumes approximately twenty times more energy, indicating that efficiency is a function of more than just number of model parameters. These performance differences likely reflect variation in model architecture, instruction fine-tuning procedures, and model-prompt compatibility rather than parameter count alone. Gemma 3 is the most energy-efficient model we tested, and performs reasonably well across tasks.

\begin{figure}[H]
\centering
\includegraphics[width=\linewidth]{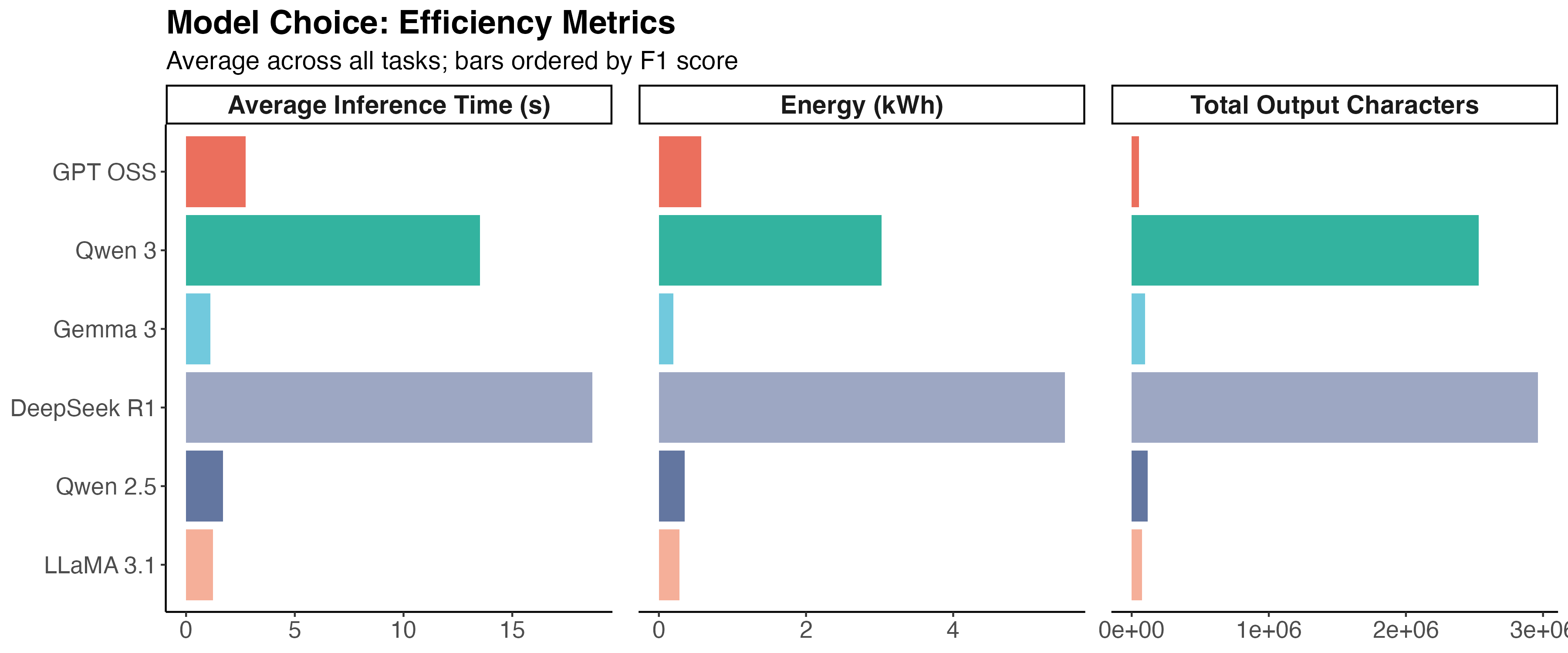}
\caption{Model choice performance and efficiency metrics. Panels compare average F1 score, inference time, energy consumption, and output length for each model, aggregated across tasks.}
\label{fig:model-choice-metrics}
\end{figure}

Further disaggregating efficiency metrics in Figure \ref{fig:model-choice-metrics} reveals a more nuanced picture. While models that consume more energy also tend to have longer inference times, the correspondence between energy consumption and the other efficiency metrics is not perfect, underscoring the importance of considering multiple dimensions of efficiency rather than relying on any single measure.

DeepSeek R1 takes the longest time to respond to a query by a considerable margin, consistent with its high energy consumption. GPT OSS produces the shortest responses on average but falls in the middle of the pack in terms of both inference time and energy consumption, despite being the smallest model tested.

Observing total output characters, DeepSeek R1 and Qwen 3 appear least willing to obey response constraints held in the prompt (i.e. \textit{``Respond only with one of the following options: $<$option1$>$, or $<$option2$>$, or $<$option3$>$. Return your response in JSON format, with the key `response'"}), as they generate considerably more characters of output compared to other models.  

Gemma 3 is the most energy-efficient model and has the shortest inference time, though it produces longer responses than GPT OSS and Qwen 2.5---illustrating that which model is ``most efficient'' depends on the metric considered. Though GPT OSS and Qwen 3 are the most performant models, Qwen 3's strong performance comes with relatively high energy consumption, long responses, and the second longest inference time after DeepSeek R1.

\begin{figure}[H]
\centering
\includegraphics[width=\linewidth]{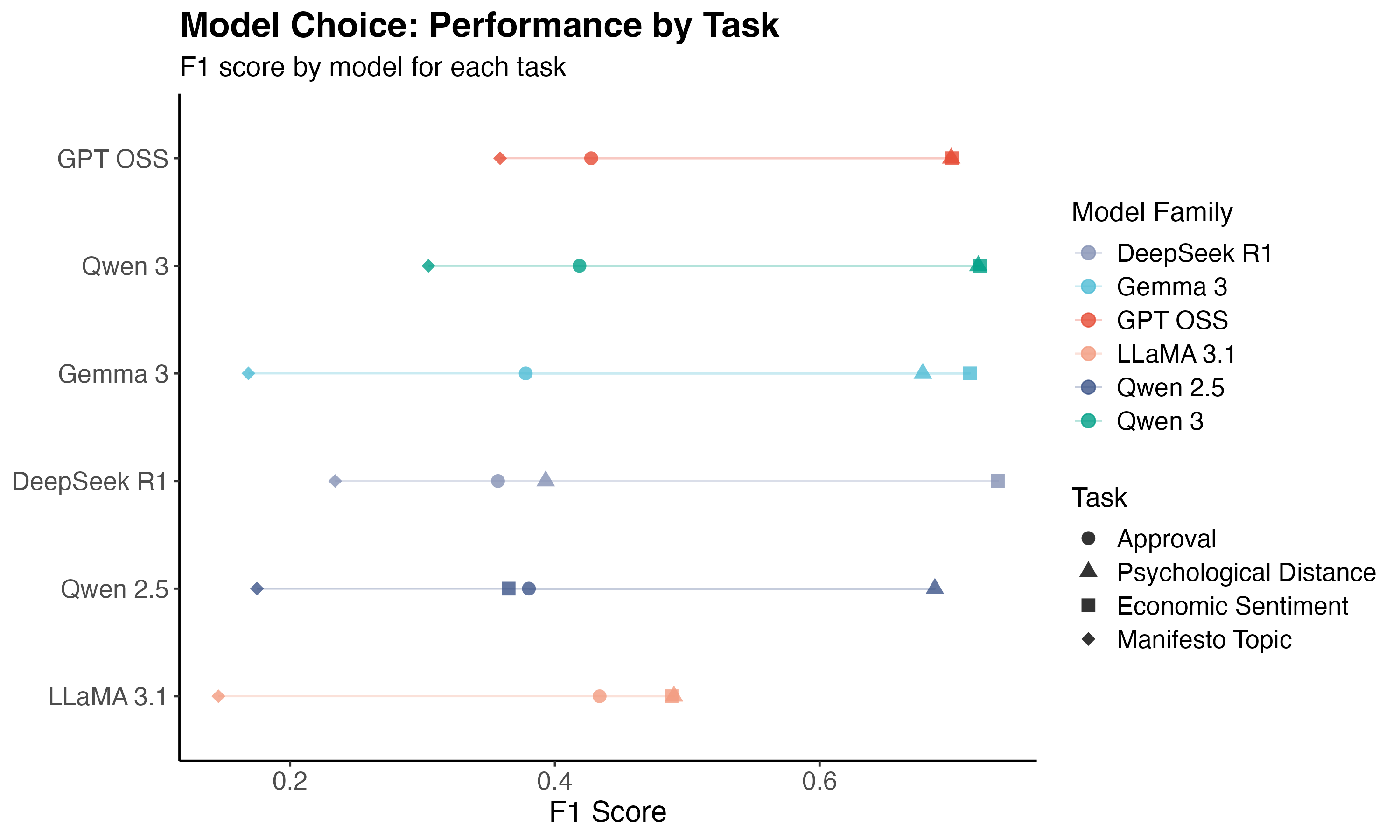}
\caption{Model choice performance by task. Task-level F1 scores are shown for each model across the four annotation tasks.}
\label{fig:model-choice-by-task}
\end{figure}

Figure \ref{fig:model-choice-by-task} shows the performance spread for each model by task. Task leadership is distributed across models: LLaMA 3.1 leads on approval, Qwen 3 on psychological distance, DeepSeek R1 on economic sentiment, and GPT OSS on the manifesto topic task. No single model dominates across all tasks, reinforcing the centrality of task--model interaction effects. Even the strongest overall performers show considerable variation: Qwen 3 achieves F1 scores in excess of 0.7 on the psychological distance and economic sentiment tasks but performs considerably worse on the two remaining tasks. This indicates that no model can simply be used out-of-the-box without first establishing that it can approximate human performance on a given annotation task. Further sensitivity checks should also be conducted, as precision may be more relevant than recall in some cases, depending on application, or vice versa.

Our initial determination around model choice is that, while certain models perform better than others on average, the variation is too task-dependent to arrive at any universal best practice.  Furthermore, greater consideration should be paid to model efficiency, as some model choices will be more suitable to certain tasks based on the volume of data to be annotated and resource constraints. Scholars should reflect carefully on whether slightly increased performance for their task is worth greater requirements in terms of time and energy use.

\subsection{Model Size}

A persistent intuition in the field is that bigger models are better: more parameters should mean more capacity to follow complex annotation instructions. We test this by comparing multiple size variants within three model families.

\begin{figure}[H]
\centering
\includegraphics[width=\linewidth]{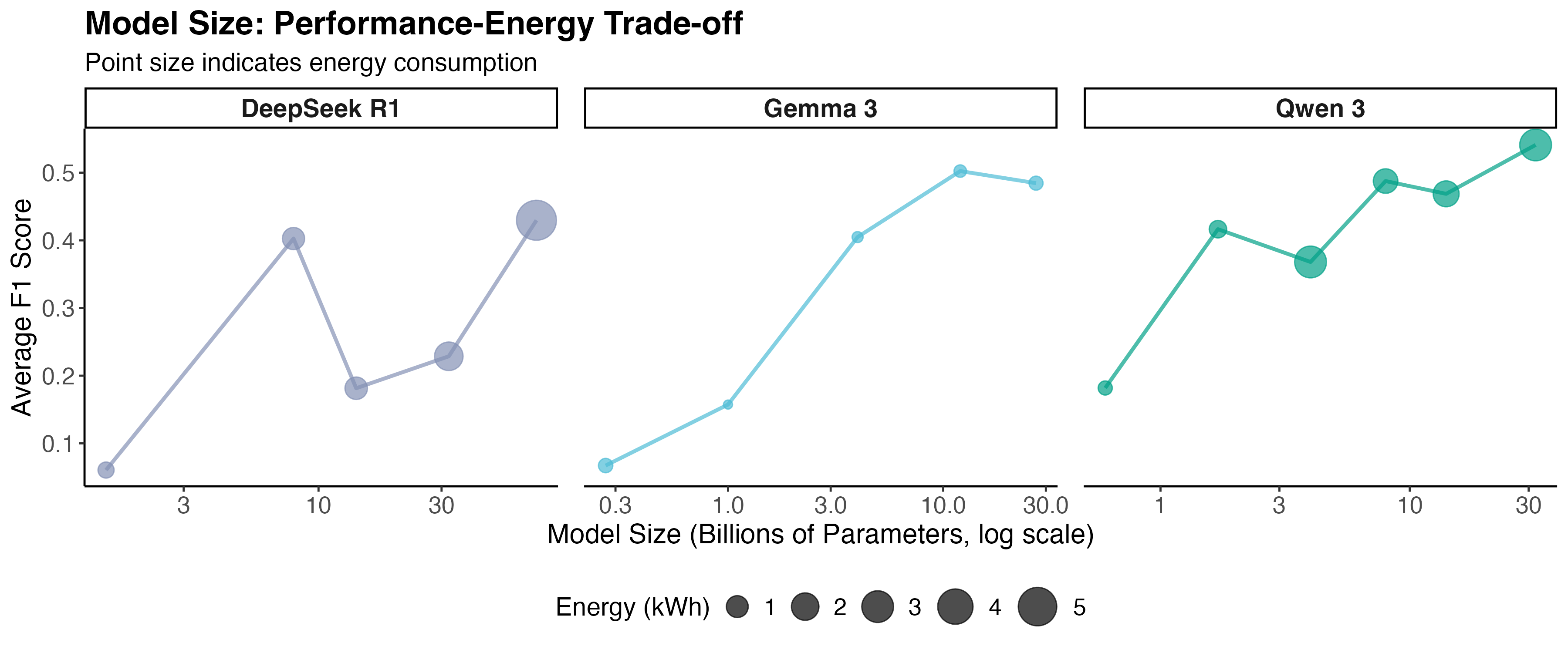}
\caption{Model size performance--energy trade-off within model families. Panels show average F1 score against model size for Gemma 3, DeepSeek R1, and Qwen 3 variants, with point size indicating energy consumption.}
\label{fig:model-size-tradeoff}
\end{figure}

The Gemma 3 family exhibits relatively consistent performance gains for each increase in model size in Figure \ref{fig:model-size-tradeoff}, though this seems to level off after a certain threshold. Curiously, the DeepSeek R1 family peaks in performance at 8B parameters, before dropping off sharply, only starting to recover again at the largest 70B parameter variant. The performance for the Qwen 3 family appears to alternate between increasing and decreasing with each increase in size, though performance trends upwards overall. 

\begin{figure}[H]
\centering
\includegraphics[width=\linewidth]{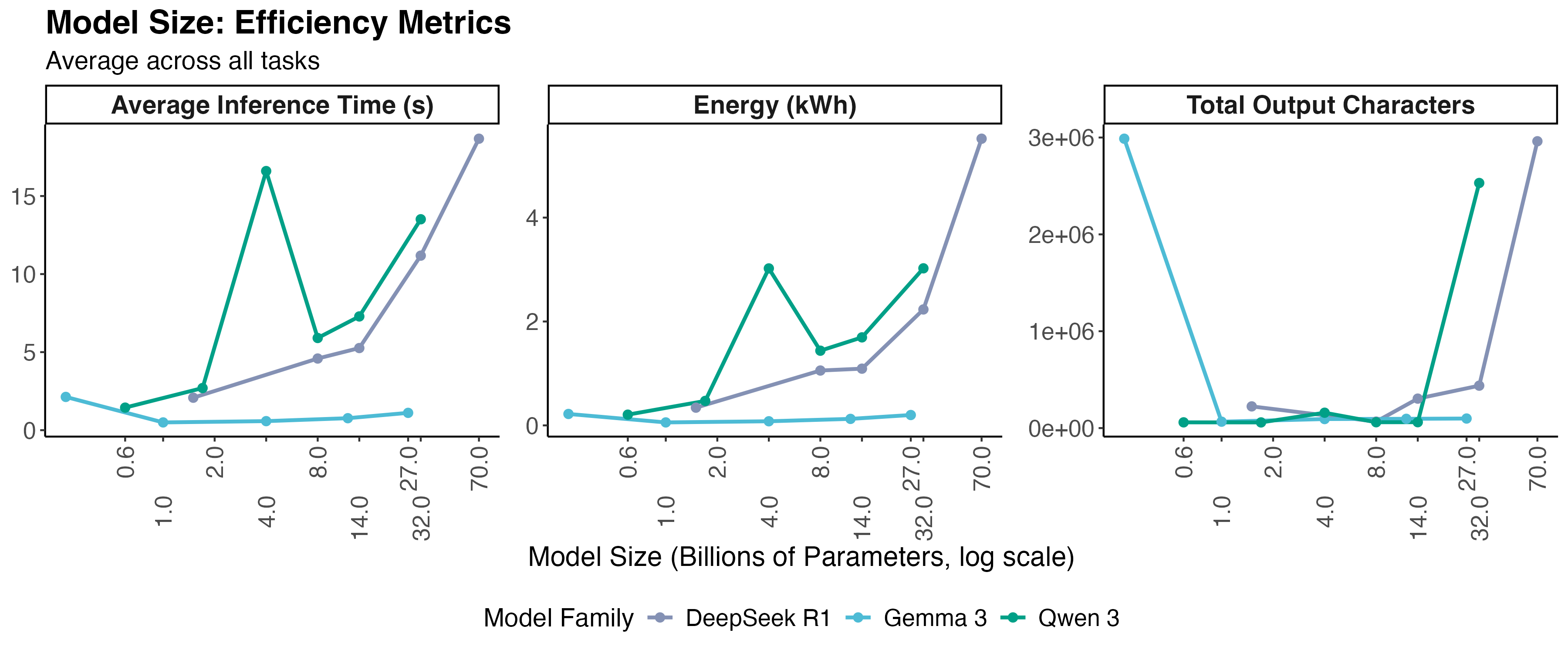}
\caption{Model size efficiency metrics within model families. Panels compare average inference time, energy consumption, and output length across Gemma 3, DeepSeek R1, and Qwen 3 size variants.}
\label{fig:model-size-metrics}
\end{figure}

Each of our efficiency metrics seems to track the others relatively well within model families but not across families, according to Figure \ref{fig:model-size-metrics}. This suggests that size often covaries with efficiency within a family, but the relationship is not strictly monotonic and includes notable exceptions. Other factors remain more important overall, as the largest variant of the Gemma 3 family is less demanding than even the smallest variants of the other families tested, despite having orders of magnitude more parameters. Notably, the smallest Gemma 3 variant (270M) and the largest variants of DeepSeek R1 (70B) and Qwen 3 (32B) exhibit weak compliance with the requested output format, generating substantially longer responses that drive up both energy consumption and inference time.

In contrast to DeepSeek R1's exponential growth in resource requirements, the increase for the Gemma 3 family appears quite modest as model size scales up. The 4B parameter version of Qwen 3 is significantly more verbose than other variants from the same family, and slower and more energy-intensive as a consequence. 

\begin{figure}[H]
\centering
\includegraphics[width=\linewidth]{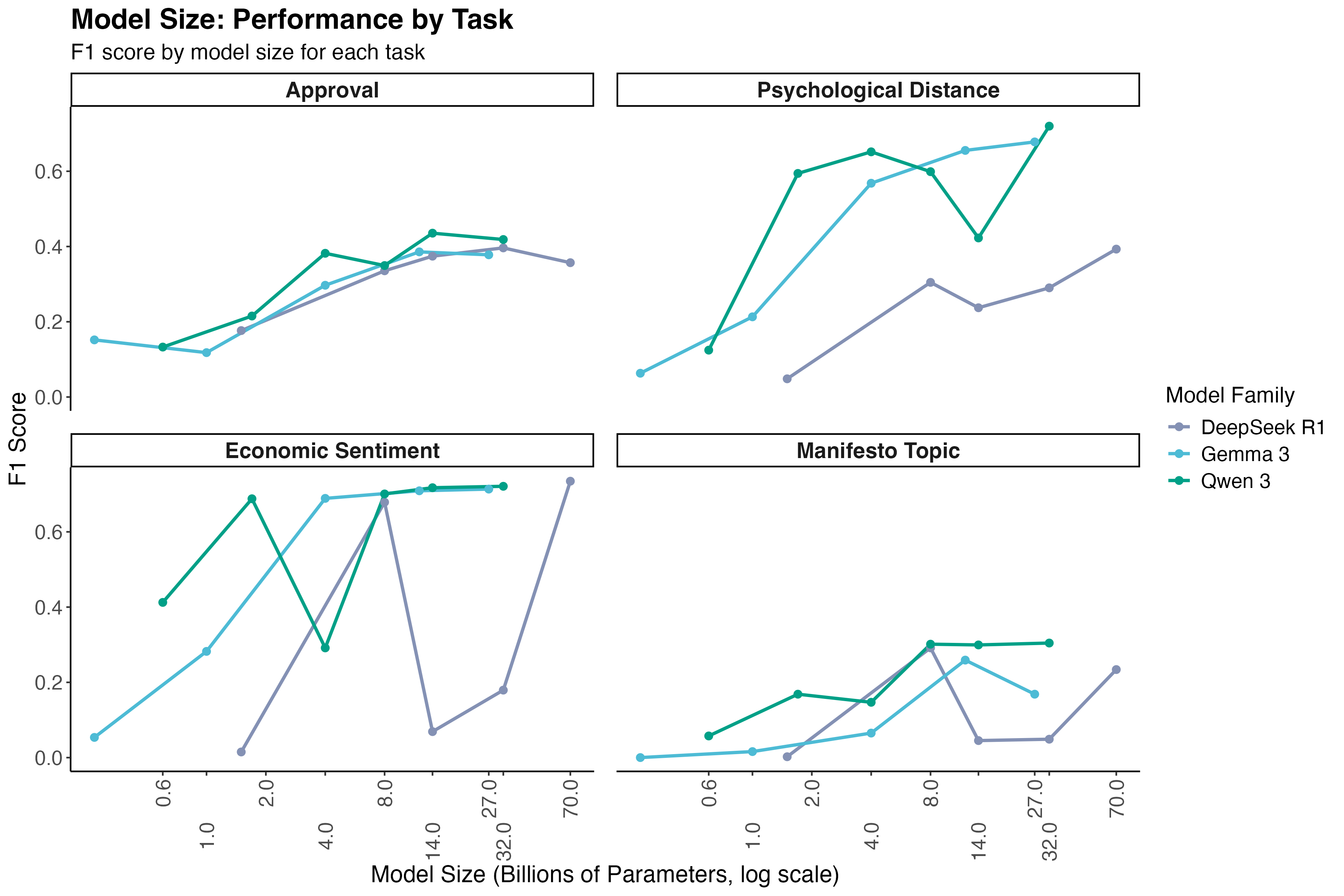}
\caption{Model size performance by task. Panels show task-level F1 scores for model variants within the Gemma 3, DeepSeek R1, and Qwen 3 families.}
\label{fig:model-size-by-task}
\end{figure}

Breaking down performance by task in Figure \ref{fig:model-size-by-task}, we see that the spike in performance for the 8B parameter variant of DeepSeek R1 (and subsequent dropoff in performance for larger variants) is driven by the economic sentiment and manifesto topic tasks, while this family's performance increases relatively linearly with size for the other two tasks. Similarly, peaks and valleys in performance are observed for the Qwen 3 family depending on task.

Together, these findings disrupt both rules-of-thumb that bigger is better and that bigger is more resource-intensive. We find that large models may be considerably less demanding than small models of a different family. We also find that resource use does not increase with model size at the same rate between families, nor does it necessarily increase consistently within a given family. 

Even holding model family constant, increasing size may decrease performance and/or increase resource consumption. This further reinforces the idea that there are no universal best practices and that LLM decisions should be empirically justified based on the specific application.

These non-monotonic patterns likely reflect, in part, differences in how model families produce smaller variants. Model developers typically reduce model size through \textit{knowledge distillation}, in which a smaller ``student'' model is trained to reproduce the outputs of a larger ``teacher'' model, or through \textit{pruning}, in which layers, neurons, or attention heads are removed from a larger model, often with subsequent retraining to recover accuracy. In practice, the approach varies across and even within model families.

The Gemma 3 and Qwen 3 families both use knowledge distillation to produce smaller variants from a common architecture \citep{team_gemma_2025,yang_qwen3_2025}, which should in principle yield consistent behaviour across sizes. Gemma 3's relatively monotonic scaling is consistent with this expectation. Qwen 3's oscillating performance across sizes may reflect varying distillation fidelity at different compression ratios. The DeepSeek R1 family takes a different approach entirely: rather than distilling from a single architecture, it fine-tunes pre-existing, independently trained base models from other families---Qwen 2.5 at 1.5B, 7B, 14B, and 32B parameters, and Llama at 8B and 70B---using reasoning samples generated by the full 671B parameter teacher \citep{deepseek-ai_deepseek-r1_2025}. The 8B performance spike thus compares a fundamentally different base architecture (Llama 3.1) to its nominal family members (Qwen 2.5). Notably, this architectural inconsistency only became apparent after investigating the anomalous results; nothing in the Ollama naming convention signals that size variants use different base architectures. This illustrates a broader challenge: pipeline choices can interact with easily overlooked model properties in ways that are difficult to anticipate without both empirical validation and close reading of technical documentation.

\subsection{Zero-Shot vs. Few-Shot Learning}

The effect of providing labelled examples, i.e. the distinction between zero-shot and few-shot prompting, is one of the most frequently cited recommendations in LLM annotation guidelines \citep{tornberg_best_2024}. Our results, however, reveal a substantially more nuanced picture than the conventional wisdom that few-shot learning uniformly improves performance.

\begin{figure}[H]
\centering
\includegraphics[width=\linewidth]{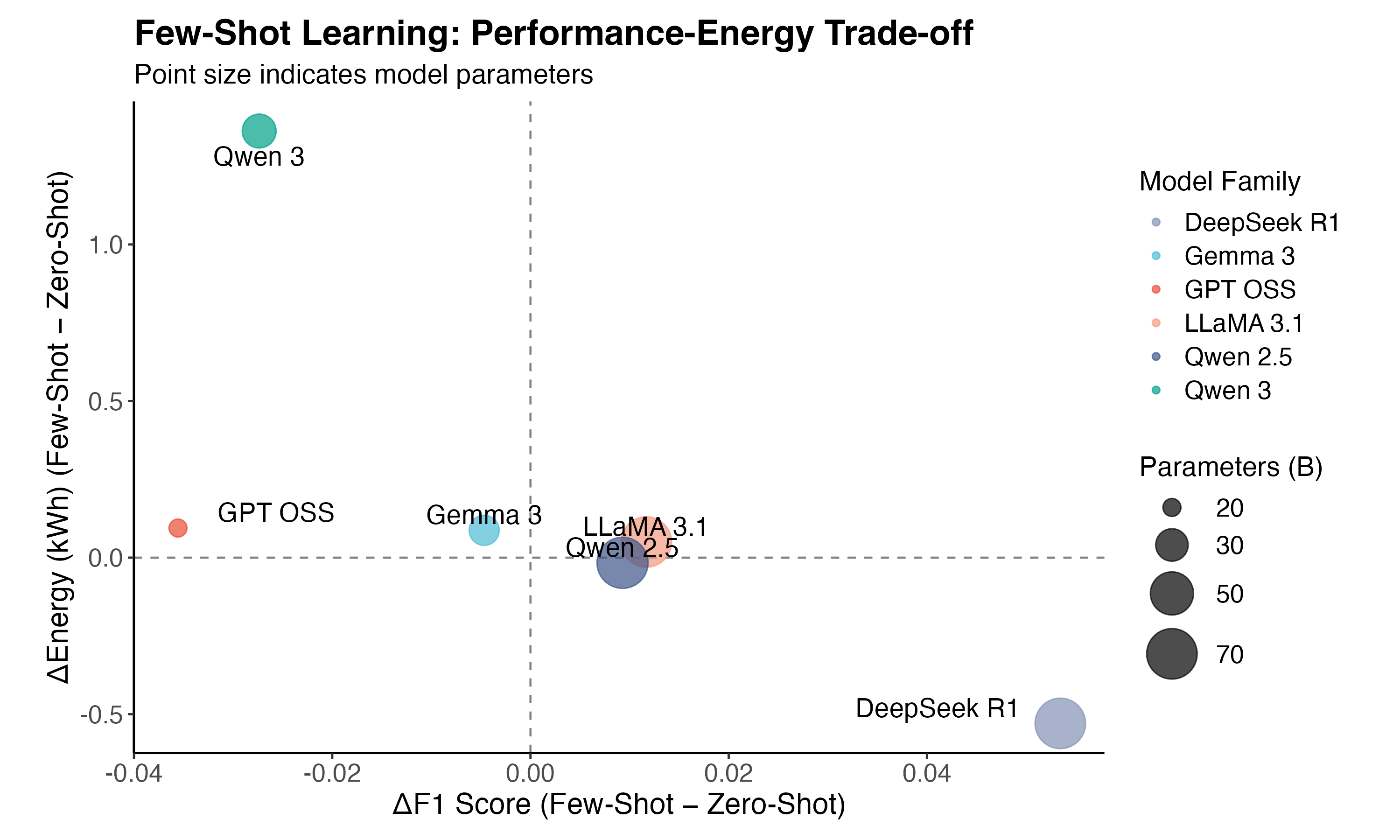}
\caption{Few-shot performance--energy trade-off. Points show the change in average F1 score and energy consumption when moving from zero-shot to few-shot prompting for each model, with point size indicating model parameters.}
\label{fig:few-shot-tradeoff}
\end{figure}

Figure~\ref{fig:few-shot-tradeoff} plots the change in average F1 against the change in energy consumption when moving from zero-shot to few-shot conditions, revealing the tradeoff involved for each model. The direction of the effect is strikingly model-dependent. DeepSeek~R1 is the clearest beneficiary, improving by approximately 0.05 F1 points under few-shot prompting while also consuming less energy. LLaMA~3.1 and Qwen~2.5 show marginal gains, while Gemma~3 is essentially unchanged. Most notably, GPT~OSS and Qwen~3---the two strongest zero-shot performers---\textit{degrade} under few-shot conditions, with GPT~OSS dropping by approximately 0.04 and Qwen~3 by approximately 0.03 F1 points. Qwen~3 also consumes significantly more energy under few-shot conditions, placing it in the worst-case quadrant of the tradeoff. This represents a cautionary finding for researchers who assume that providing examples is uniformly beneficial: the models that perform best without examples may actually be harmed by their inclusion.

\begin{figure}[H]
\centering
\includegraphics[width=\linewidth]{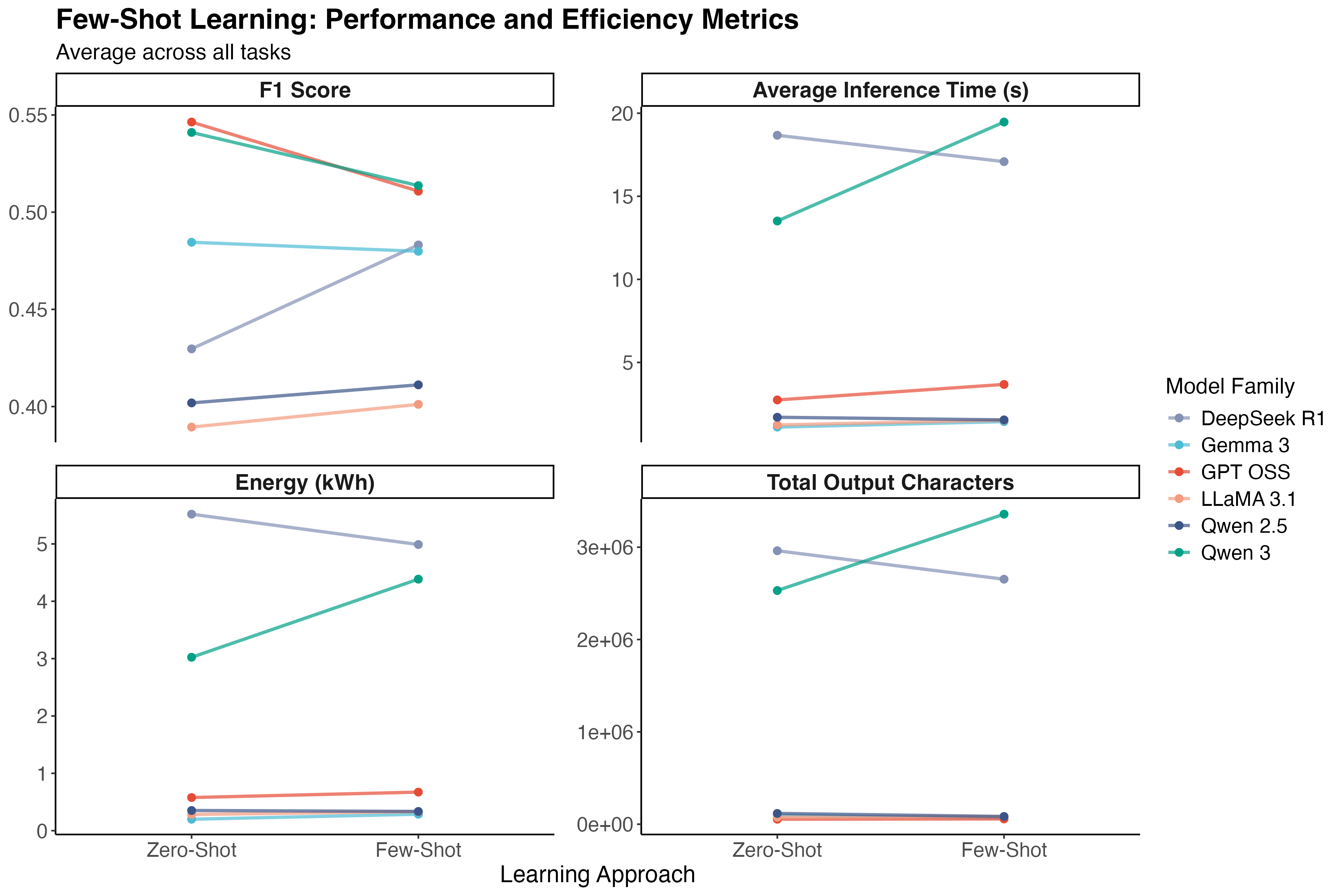}
\caption{Few-shot performance and efficiency metrics. Panels compare average F1 score, inference time, energy consumption, and output length under zero-shot and few-shot prompting across models.}
\label{fig:few-shot-metrics}
\end{figure}

The efficiency implications of few-shot prompting are also more complex than commonly assumed. Figure~\ref{fig:few-shot-metrics} shows that DeepSeek~R1 becomes substantially \textit{more} energy-efficient under few-shot conditions, while Qwen~3 becomes considerably \textit{less} efficient. The remaining models show relatively modest changes. Gemma~3 remains the most energy-efficient model under both conditions.

\begin{figure}[H]
\centering
\includegraphics[width=\linewidth]{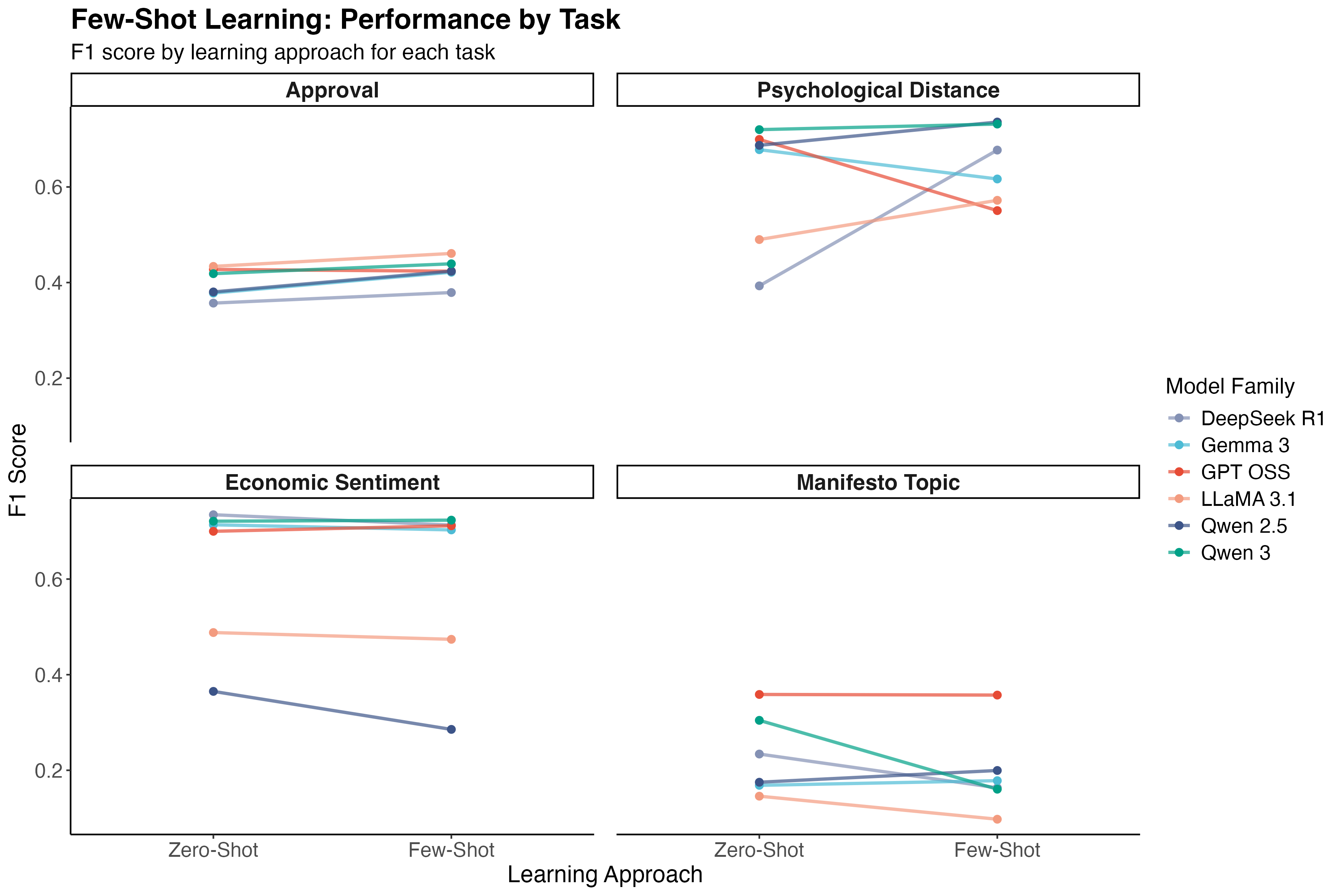}
\caption{Few-shot performance by task. Panels show task-level F1 scores for each model under zero-shot and few-shot prompting.}
\label{fig:few-shot-by-task}
\end{figure}

Decomposing performance by task in Figure~\ref{fig:few-shot-by-task} reveals that the aggregate patterns mask considerable task-level heterogeneity. The psychological distance task shows the most widespread benefit from few-shot learning, with DeepSeek~R1 exhibiting a particularly dramatic gain (from approximately 0.39 zero-shot F1 to 0.68 under few-shot conditions). This is consistent with the task's nested, multi-step structure, which may benefit from worked examples that clarify the annotation logic. However, even here the effect is not universal: GPT~OSS degrades under few-shot conditions on this task. By contrast, the approval task is largely insensitive to learning approach, with most models performing similarly under both conditions, suggesting that this relatively straightforward Likert-scale task is well-served by instructions alone.

The economic sentiment task shows most models maintaining roughly similar performance under both conditions. On the manifesto topic task, results are mixed: several models show modest declines, possibly because the eight-category scheme with accompanying definitions already provides substantial in-context information, and additional examples add noise or prompt-length overhead.

Together, these findings challenge the blanket recommendation to use few-shot prompting for annotation tasks. The decision to provide examples is itself a researcher degree of freedom with model- and task-dependent consequences. For some model--task combinations, few-shot prompting delivers meaningful improvement; for others, it degrades performance or simply adds computational overhead. Researchers should empirically test both approaches on their specific model and task before committing to either.

\subsection{Prompt Style}

Prompt engineering---iteratively modifying instructions to improve model output---is perhaps the most accessible lever available to researchers deploying LLMs as annotators, and the one most susceptible to undisclosed analytical flexibility. Without a held-out test set, each modification risks overfitting to the validation sample rather than improving genuine annotation ability. Our interest here is not only whether two prominent techniques help in themselves, but what they reveal about prompt engineering as a practice. We examine persona prompting (framing the model as a domain expert) and chain-of-thought prompting (instructing the model to reason step-by-step) as case studies: both are widely recommended, theoretically motivated, and easy for researchers to adopt without formal validation. We do not claim to evaluate prompt engineering exhaustively; rather, we use these two well-motivated interventions to illustrate the risks posed by iterative prompt modification without a separate test set. Their unstable effects in our experiments suggest that prompt engineering should be treated as a consequential tuning decision requiring held-out evaluation, rather than as a source of portable best practices.

\begin{figure}[H]
\centering
\includegraphics[width=\linewidth]{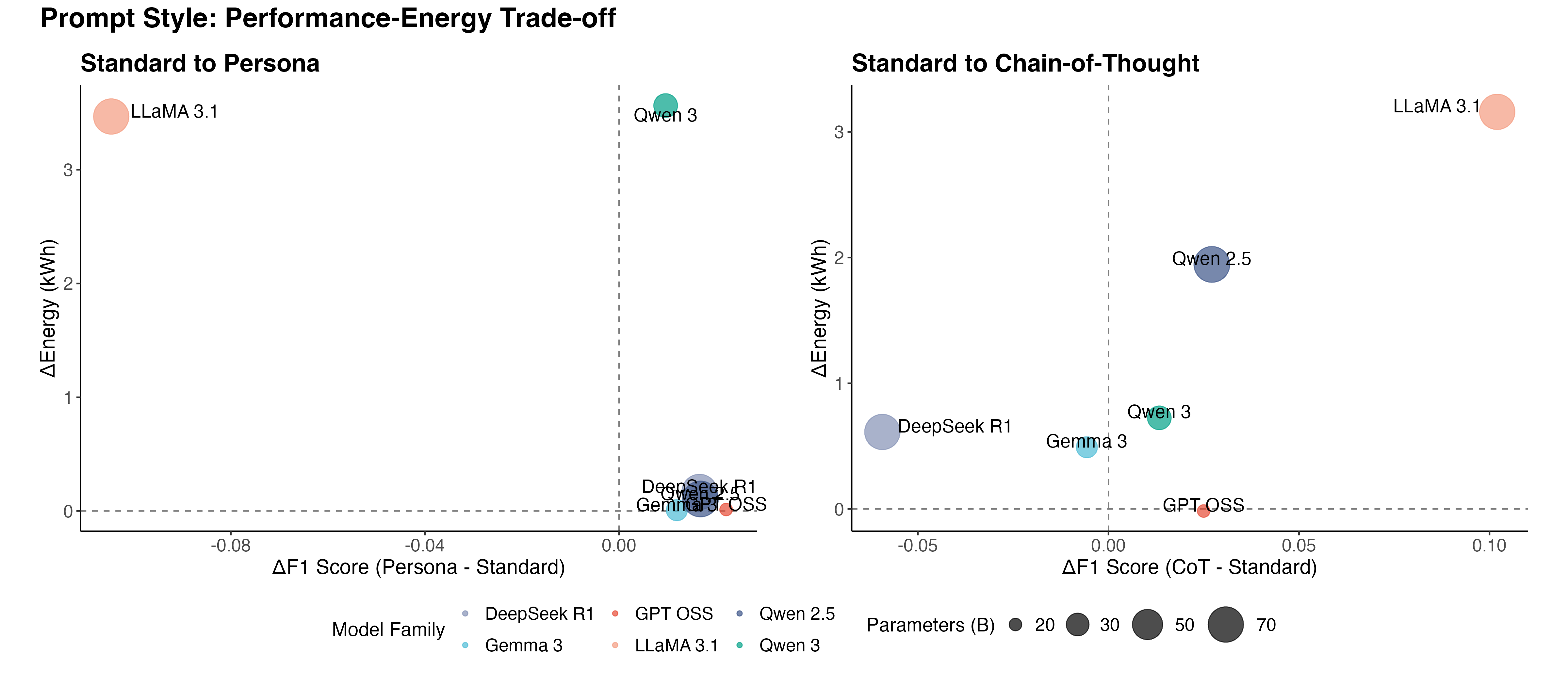}
\caption{Prompt style performance--energy trade-off. The left panel shows the change in average F1 score and energy consumption from standard to persona prompting, and the right panel shows the same comparison for standard to chain-of-thought prompting.}
\label{fig:prompt-style-tradeoff}
\end{figure}

Figure~\ref{fig:prompt-style-tradeoff} plots the change in F1 against the change in energy consumption when moving from standard prompting to persona (left panel) and CoT (right panel) conditions. The ideal outcome---improved performance at low additional cost---would place a model in the lower-right quadrant. Instead, most models cluster in the upper half (energy increases) with F1 changes near zero or negative, and the aggregate picture is once again one of inconsistency. Neither persona nor CoT prompting reliably improves over the standard baseline. The most dramatic effect is negative: Llama~3.1 experiences a severe collapse under persona prompting, with average F1 dropping from approximately 0.40 to below 0.30. By contrast, GPT~OSS improves modestly under both persona and CoT conditions at relatively low energy cost, owing to its small active parameter count under its mixture-of-experts architecture. Qwen~3 performance is largely invariant to prompt style but incurs a substantial energy penalty under persona conditions. DeepSeek~R1 improves under persona prompting but occupies the worst-case position under CoT: increased energy consumption paired with \textit{decreased} performance. Llama~3.1 shows a substantial F1 gain under CoT but at the highest energy cost of any model, representing a high-cost, high-reward tradeoff. The net effect of either technique, averaged across models, is close to zero, with variance that far exceeds the mean.

\begin{figure}[H]
\centering
\includegraphics[width=\linewidth]{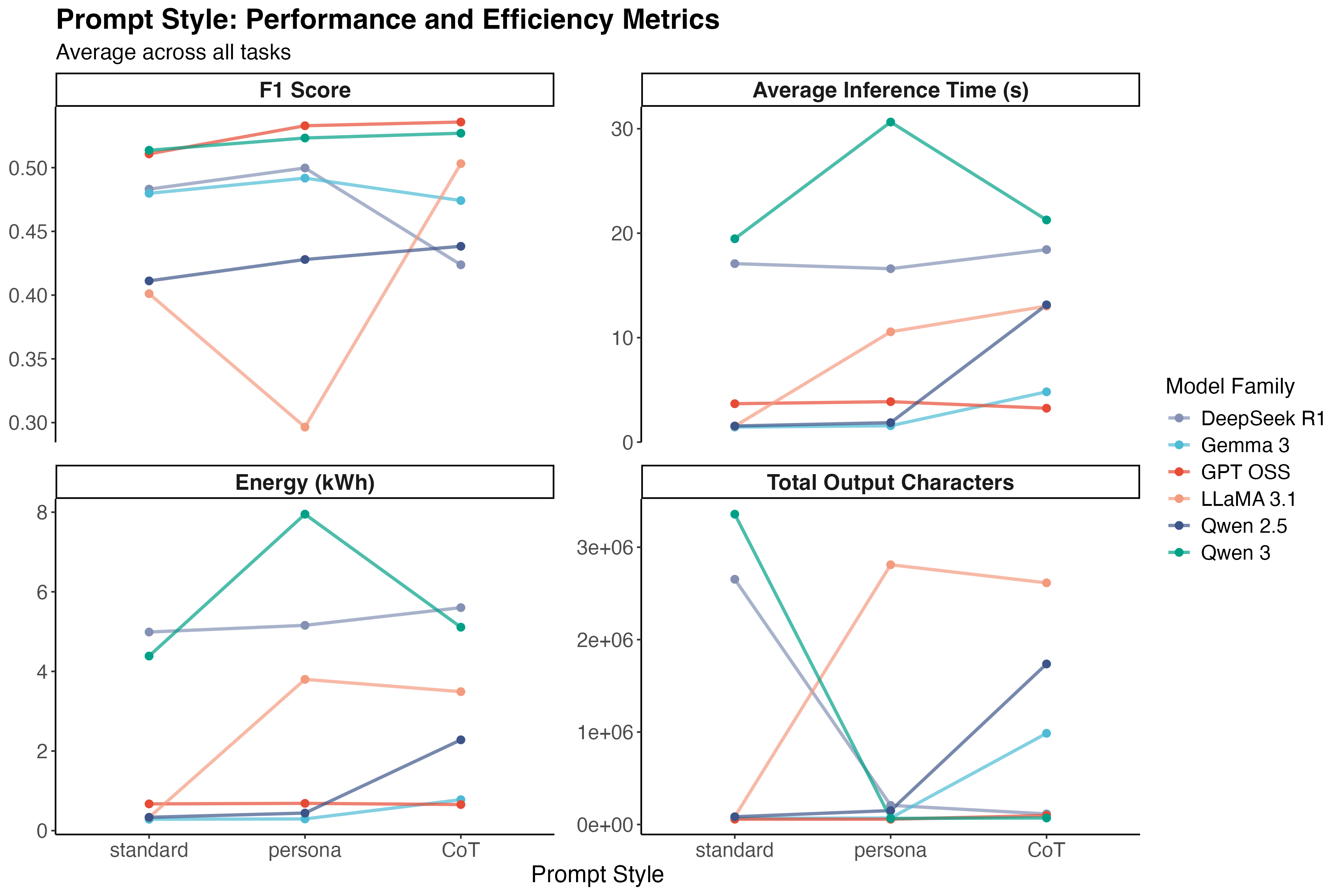}
\caption{Prompt style performance and efficiency metrics. Panels compare average F1 score, inference time, energy consumption, and output length across standard, persona, and chain-of-thought prompting.}
\label{fig:prompt-style-metrics}
\end{figure}

The decomposed efficiency metrics in Figure~\ref{fig:prompt-style-metrics} reveal a more nuanced pattern than a simple ``CoT costs more'' narrative. GPT~OSS spikes in response length under both persona and CoT conditions, with only marginal effects on performance. By contrast, response length for Qwen~3 and DeepSeek~R1 \textit{plummets} under persona and CoT conditions, suggesting that these reasoning models produce more focused outputs when given explicit reasoning instructions but not necessarily better ones. The energy picture does not track response length: Qwen~3 exhibits a dramatic spike in energy consumption under persona prompting despite producing shorter responses, while energy costs for the other models vary in both direction and magnitude across prompt styles.

\begin{figure}[H]
\centering
\includegraphics[width=\linewidth]{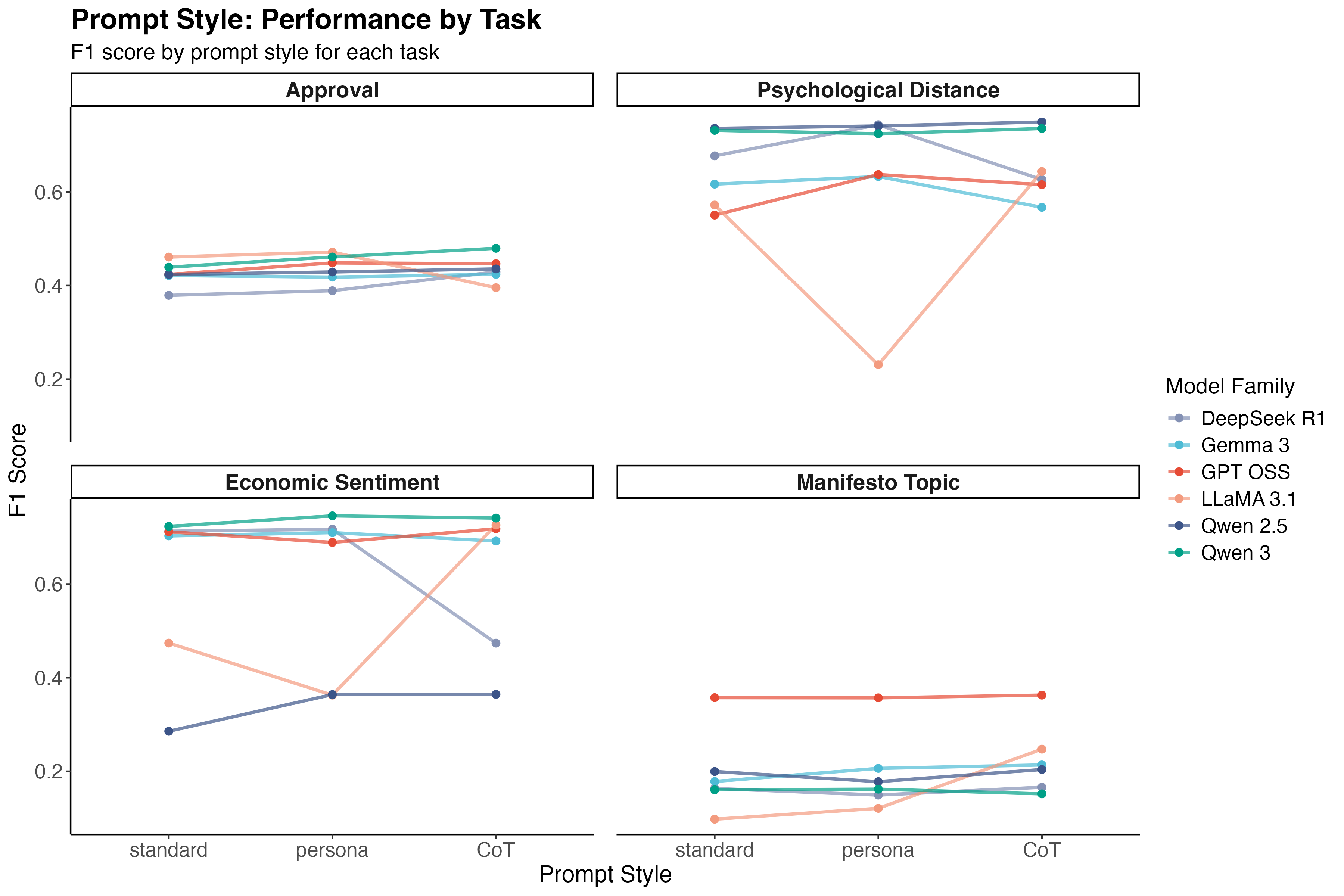}
\caption{Prompt style performance by task. Panels show task-level F1 scores for each model across standard, persona, and chain-of-thought prompting.}
\label{fig:prompt-style-by-task}
\end{figure}

The task-level decomposition in Figure~\ref{fig:prompt-style-by-task} helps explain the aggregate null result. The approval task is largely insensitive to prompt style for most models, though GPT~OSS notably degrades under CoT, suggesting that for simple tasks the additional reasoning can actively hurt performance. The psychological distance task shows more divergent effects: Llama~3.1 improves substantially under CoT (from 0.57 to 0.64 F1) but \textit{collapses} under persona prompting (from 0.57 to 0.23 F1), illustrating how the same model can respond in opposite directions to different prompting interventions on the same task. Economic sentiment, despite being a relatively straightforward binary task, sees performance \textit{degrade} under CoT for DeepSeek~R1 (from 0.71 to 0.47 F1), possibly because extended reasoning introduces second guessing on cases where the initial judgement was correct. The manifesto topic task shows minimal systematic differences across prompt styles.

The practical implications of our findings are clear. Our evidence does not support the routine use of persona or chain-of-thought prompting for LLM-based text annotation. The performance effects of both techniques are erratic, showing positive effects for some model--task combinations and negative for others. The efficiency costs of these techniques are similarly model-dependent: some models incur substantial energy increases while others show modest or even reduced costs, but these efficiency shifts do not reliably correspond to performance improvements. Researchers who adopt these techniques without sufficient attention to task-specific validation exercises risk both degrading their measurements and incurring unnecessary computational costs.

\section{A Validation-First Framework for LLM Annotation}\label{sec:validation-framework}

The preceding results paint a consistent picture: the effect of any single pipeline choice on annotation performance depends on its interaction with the model, the task, and other pipeline choices. This pervasive task-dependence means that there are no universal ``best practices'' for LLM annotation, but this does not mean that anything goes. In this section, we distil our findings into a practical framework for researchers designing LLM annotation pipelines.

\subsection{A Validation Workflow}

Any LLM annotation pipeline should be pilot-tested against a human-coded validation sample before being deployed at scale, echoing the third principle of \citet{grimmer_text_2013}. Our results imply a specific order for this validation process:

\begin{enumerate}
    \item \textbf{Define efficiency constraints first.} Cross-family efficiency differences dominate within-family differences: Gemma~3 at 27B parameters is less resource-intensive than the smallest DeepSeek R1 variant at 1.5B parameters (Figure \ref{fig:model-size-metrics}). Shortlist two to three model families that fit your hardware, budget, and timeline before evaluating performance.
    \item \textbf{Run candidate models zero-shot with standard prompting on a human-coded validation sample.} Model choice is the highest-impact decision, with spreads exceeding 0.15 F1 points across models (Figure \ref{fig:model-choice-tradeoff}). Standard zero-shot prompting is the cheapest baseline to evaluate; a few hundred documents stratified across categories is typically going to be sufficient.
    \item \textbf{Select the best-performing model on your task.} Do not rely on published benchmarks. As we demonstrate, task--model interactions dominate main effects. Task leadership is distributed across four different models, with no single model dominating (Figure \ref{fig:model-choice-by-task}).
    \item \textbf{Test few-shot on your selected model.} The few-shot effect is model-dependent (Figure \ref{fig:few-shot-tradeoff}): strong zero-shot performers (GPT~OSS, Qwen~3) may actually degrade, while weak zero-shot performers can improve dramatically (DeepSeek R1 on psychological distance: 0.39 $\rightarrow$ 0.68 F1). If zero-shot performance is already satisfactory, skip this step.
    \item \textbf{Only test prompt styles if performance remains unsatisfactory.} Prompt engineering is the lowest-impact decision, with erratic performance effects and model-dependent efficiency costs (Figure \ref{fig:prompt-style-tradeoff}, Figure \ref{fig:prompt-style-metrics}).
\end{enumerate}

This sequence of decisions is methodologically sensible, but it also creates a risk of overfitting. Each pipeline decision---model selection, prompt wording, codebook design---functions as an adjustable hyperparameter. Researchers who iteratively modify prompts or codebook language to improve performance on a validation sample are, in effect, tuning the pipeline to a fixed set of documents rather than to the underlying task. If these modifications deviate from the instructions given to human annotators, the resulting comparison between human and LLM performance is no longer on equal footing. Our finding that performance varies unpredictably across tasks and configurations (Figure \ref{fig:model-choice-by-task}, Figure \ref{fig:prompt-style-by-task}) reinforces this concern: gains observed on one sample may not generalise. The design reported here was constructed to minimise this risk: task definitions, response options, and examples were frozen before benchmarking, and prompt-style manipulations were specified in advance as common template variants rather than tuned on the basis of model performance.

The standard remedy in machine learning is a three-way split into training, validation, and test sets, where the test set is withheld until all development decisions are finalised \citep[see e.g.][]{goodfellow_deep_2016}. We recommend the same discipline for LLM annotation: researchers who undertake iterative prompt engineering or codebook revision should maintain a held-out test set and report final performance on it rather than on the validation sample used to guide design choices. Under this design, few-shot examples should be drawn from a training partition, pipeline variants compared on a validation partition, and the final workflow reported once on a held-out test partition. Alternatively, researchers can sidestep this problem by providing the model with the same codebook distributed to human annotators, without further modification. This preserves the two-way split, avoids the degrees of freedom introduced by prompt engineering, and ensures a direct comparison between human and LLM performance on identical instructions.

\subsection{Open Questions: Task Type, Architecture, and Efficiency}

The preceding workflow treats all tasks alike, but our per-task decompositions suggest that task characteristics may predict which pipeline choices matter. For instance, the simple approval task is largely insensitive to both prompt style and few-shot learning, while the complex psychological distance task shows more widespread few-shot benefits. Whether these patterns generalise beyond our four tasks is an open question: our task set is too narrow to support confident heuristics about task-type effects, and we flag this as a priority for future work with a broader and more diverse set of annotation tasks.

Two practical observations do emerge with more generality from our results. First, \textit{model architecture determines efficiency more than parameter count}. When choosing models under resource constraints, researchers should compare efficiency \emph{across} families rather than \emph{within} them. Gemma~3 at 27B parameters is less resource-intensive than the smallest DeepSeek R1 variant (Figure \ref{fig:model-size-metrics}). A model half the size from a different family can be more expensive to run.

Second, \textit{reasoning models may respond differently to prompting interventions}. Models trained with reinforcement learning to produce intermediate reasoning steps (Qwen~3, GPT~OSS, DeepSeek~R1) might be expected to benefit less from chain-of-thought prompting, since explicit CoT instructions are partially redundant with built-in reasoning capabilities \citep[see e.g.][]{sprague_cot_2025}. Our results offer partial support: DeepSeek~R1 degrades under CoT for some tasks, and Qwen~3 is largely invariant to prompt style (Figure \ref{fig:prompt-style-by-task}). However, GPT~OSS improves under CoT, and the few-shot results do not split cleanly along reasoning lines either. The reasoning distinction is therefore a useful heuristic---researchers should be cautious about adding CoT prompts to reasoning models---but it does not override the task--model interactions that dominate our findings.

\subsection{Reporting Standards}

A recurring theme of our results is that pipeline choices interact in ways that are difficult to predict from first principles. Model rankings shift across tasks (Figure \ref{fig:model-choice-by-task}), the effect of few-shot examples and prompt style varies by model and task (Figure \ref{fig:few-shot-by-task}, Figure \ref{fig:prompt-style-by-task}). These interactions mean that omitting pipeline details from a published study makes it impossible for readers to assess whether the reported performance would hold under alternative, equally defensible configurations. In adjacent fields, initiatives such as model cards \citep{mitchell_model_2019} and datasheets for datasets \citep{gebru_datasheets_2021} have established norms around structured transparency; LLM-based annotation in the social sciences would benefit from analogous standards. We recommend that studies using LLM-based annotation report, at minimum, the following:

\begin{enumerate}
    \item \textbf{Model identity:} the model name, version string, and parameter count. Our model choice experiments show performance spreads exceeding 0.15 macro F1 across models of comparable size (Figure \ref{fig:model-choice-tradeoff}); model identity is therefore essential for interpreting any reported result.
    \item \textbf{Quantisation:} the quantisation method and level (e.g., GGUF Q4\_K\_M). All models in our study were evaluated at identical quantisation; varying this parameter introduces a further dimension of analytical flexibility that should be made explicit.
    \item \textbf{Prompt text:} the full prompt or a representative template, including system instructions. Our prompt style results demonstrate that seemingly minor changes to prompt wording (e.g., adding a persona prefix or chain-of-thought instruction) can shift F1 by several points in either direction with unpredictable effects on computational cost (Figure \ref{fig:prompt-style-tradeoff}, Figure \ref{fig:prompt-style-metrics}).
    \item \textbf{Sampling hyperparameters:} at minimum, temperature, top\_k, and top\_p. These settings govern the stochasticity of model output and vary across models (Table~\ref{tab:model-comparison}), but can also be modified by the researcher; omitting them hinders replication.
    \item \textbf{Learning approach:} whether zero-shot, few-shot, or some other prompting was used, and if few-shot, the number and selection method of examples. Few-shot effects are strongly model- and task-dependent: some models improve substantially on certain tasks while others degrade (Figure \ref{fig:few-shot-by-task}). Without this information, neither replication nor comparison across studies is possible.
    \item \textbf{Hardware specification:} the GPU model, memory, and any relevant infrastructure details (e.g., inference framework, batch size). Our efficiency measurements are conditioned on specific hardware (NVIDIA A100 GPUs) and are likely to differ on other configurations.
    \item \textbf{Efficiency metrics:} aggregate inference time and energy consumption alongside performance metrics. Our results show that pipeline choices with negligible effects on annotation performance can produce order-of-magnitude differences in computational cost (Figure \ref{fig:prompt-style-tradeoff}, Figure \ref{fig:prompt-style-metrics}). Reporting only performance obscures these trade-offs.
\end{enumerate}

This information is necessary both for assessing the replicability of individual studies and for enabling the kind of cross-study comparisons that would allow the field to accumulate generalisable knowledge about LLM annotation pipelines.

\section{Conclusion}

This paper uses a controlled benchmark of LLM annotation pipelines to make a methodological argument about political text analysis: LLM annotation should be treated as a validation problem under researcher degrees of freedom. By varying model choice, model size, prompt style, and zero-shot versus few-shot learning across four annotation tasks, we show that implementation decisions are not mere engineering details. They are part of the measurement strategy itself.

The central result is therefore not that one model family wins overall, but that interaction effects between model, task, and pipeline configuration dominate main effects throughout our results. Task-specificity is the rule, not the exception: the best-performing model varies by task, the value of few-shot examples depends on the model and task, and prompt engineering techniques such as persona and chain-of-thought prompting yield unstable effects. Two broader lessons follow. First, no portable set of ``best practices'' can substitute for task-specific empirical validation. Second, efficiency must be treated as part of methodological design rather than as a secondary implementation concern, because computational cost varies sharply across otherwise plausible pipeline choices.

These findings motivate a validation-first framework in which researchers begin from a frozen codebook, benchmark candidate pipelines against human-coded data, and report the design choices that shaped the resulting measurements. Where the goal is direct comparison between human and LLM coders, the cleanest design is often to give the model the same codebook, including examples, that was given to human annotators. Where prompt or codebook development is iterative, a stricter train/validation/test discipline is needed: few-shot examples should be drawn from a training partition, pipeline choices compared on a validation partition, and final performance reported on a held-out test set. We also propose reporting standards that make these choices transparent and release CodeBook Studio and CodeBook Lab as open-source tools that operationalise this workflow.

Three broader implications deserve emphasis. First, our results document \emph{relative} differences among pipeline configurations; they do not establish that LLM annotation is superior to fine-tuned encoder models or other established approaches. Whether an LLM pipeline is the right tool for a given task remains an empirical question that depends on the annotation scheme, the available training data, and the computational budget. Second, model-specific rankings will inevitably date as new architectures emerge, but the methodological lesson should not: pipeline heuristics require validation rather than uncritical adoption. Third, because prompt and pipeline decisions can alter measurements in substantively meaningful ways, making those decisions explicit is a prerequisite for cumulative research rather than a matter of descriptive completeness.

Our study has several limitations. We evaluate only open-weight models; closed-source systems may exhibit different sensitivity profiles and represent an important benchmark for future work. Our four tasks, while spanning a range of text types and annotation formats, do not exhaust the diversity of political science annotation tasks. Tasks involving longer documents, more granular coding schemes, or multilingual corpora may reveal different patterns, and these tasks deserve our attention. We also report point estimates from a single run of each experimental condition rather than bootstrapped confidence intervals or repeated runs.\footnote{We considered both bootstrapped confidence intervals and multiple experimental runs to quantify sampling uncertainty and model stochasticity, but the size of the experimental grid and the associated computational cost required prioritising breadth of model and pipeline coverage.} This leaves residual uncertainty around some differences and should be addressed in future work. All models were evaluated at a single quantisation level on the same hardware, so absolute efficiency estimates may differ under other quantisation settings or computational environments. Finally, we evaluate each pipeline dimension in relative isolation; future work should explore higher-order interaction effects (e.g., CoT prompting combined with few-shot examples, or the joint effect of model size and temperature), test a broader range of models including closed-source systems, and extend the evaluation to tasks with richer annotation structures. Our model-specific results will inevitably date as new architectures emerge, but the broader lesson---that pipeline heuristics require task-specific validation rather than uncritical adoption---should not. As LLM-based annotation becomes routine in political science, the field will benefit from the same culture of systematic validation that has characterised the best work in computational text analysis.

\subsection*{Data and Code Availability}

Replication materials for this paper, including codebooks, ground-truth annotations, and experimental configuration files, are available at [replication materials URL]. The LLM annotation pipeline used to run all experiments is CodeBook Lab \citep{mclaren_codebook_lab_2026}, available at \submissionhref{https://github.com/LorcanMcLaren/codebook-lab}. Codebook definitions were created using CodeBook Studio \citep{mclaren_codebook_studio_2026}, available at \submissionhref{https://github.com/LorcanMcLaren/codebook-studio} (hosted instance: \submissionhref{https://codebook.streamlit.app}). A step-by-step tutorial covering the full workflow from codebook design through LLM benchmarking is available at \submissionhref{https://lorcanmclaren.com/codebook-tutorial.html}.

\newpage
\begingroup
\setstretch{1.0}
\printbibliography
\endgroup
\newpage
\appendix
\pagenumbering{arabic}
\counterwithin{figure}{section}
\counterwithin{table}{section}
\renewcommand\thefigure{\thesection.\arabic{figure}}
\renewcommand\thetable{\thesection.\arabic{table}}

\section{Codebooks}\label{app:codebooks}
\subsection{Approval}
General approval refers to the degree to which the speaker expresses approval of the proposal(s) being negotiated or the state of negotiations. In orientation debates, this dimension will typically be approval of the legislative proposal presented by the European Commission. Later on in the negotiation process, this dimension will typically be approval of the presidency's suggestions. If the speaker presents a most preferred position and one on which they could compromise, always code the preferred position. Assess the degree of approval expressed in the speech below.

\begin{itemize}
    \item \textbf{Question:} What level of approval is expressed in the speech?
    \item \textbf{Options:} Rate on the following scale:
    \begin{enumerate}
        \item The speaker expresses full approval.
        \item The speaker expresses more approval than disapproval.
        \item The speaker expresses a balance of approval and disapproval.
        \item The speaker expresses more disapproval than approval.
        \item The speaker expresses full disapproval.
    \end{enumerate}
    If the speaker conveys their position but the exact level of approval is hard to assess, choose 3.
    \item \textbf{Examples:}
    \begin{itemize}
        \item Code `1' (Full approval):
        \begin{itemize}
            \item ``Thank you Mr. Chairman, we congratulate the Presidency for the good compromise proposal just on the table, we fully support it and can accept it. Regarding the comitology procedure on Article 24 we're part of the declaration together with the United Kingdom, Malta and Ireland. We're ready and we accept the present proposal.''
        \end{itemize}
        \item Code `2' (More approval than disapproval):
        \begin{itemize}
            \item ``Thank you Mr. President, just one word to say that we fully support the position presented by Malta 100\%, thank you.''
        \end{itemize}
        \item Code `3' (Balance of approval and disapproval):
        \begin{itemize}
            \item ``Thank you for the floor. The Czech Republic is also very much interested in finding the good compromise. However we think we should maybe work one more month in the effort to try a full compromise, we share the opinion of the Dutch delegation and the Swedish delegation. Thank you.''
        \end{itemize}
        \item Code `4' (More disapproval than approval):
        \begin{itemize}
            \item ``I recognize the Presidency's effort in finding compromise, but I am concerned about the unleveled playing field it introduces since some Member States will exchange a lot of information while other Member States will exchange only a little information. Therefore I agree with the Dutch proposal but at the same time we would accept the compromise if it were the only way to reach an agreement.''
        \end{itemize}
        \item Code `5' (Full disapproval):
        \begin{itemize}
            \item ``Thank you, we are all currently engaged in consolidating our national economies and cutting expenses on the national budgets and therefore it seems out of touch with the economic realities with Member States that the European Commission while recommending budget cuts in Member States at the same time asks for additional monetary resources for itself.''
        \end{itemize}
    \end{itemize}
\end{itemize}
\newpage

\subsection{Psychological Distance}
Respond to items 2-3 in this section only if you select `Yes' for item 1 (Presence). Otherwise, leave these blank. You may also choose to leave item 1 (Presence) blank; this will be interpreted as a `No' response.

Respond to item 3 (Proximity) only if you select `Specific' for item 2 (Specificity).

\subsubsection{Presence}
\begin{itemize}
    \item \textbf{Question:} Does this text discuss the effects of climate change, pollution, or environmental destruction, defined as the adverse effects on natural ecosystems, human health, and the planet as a whole, caused by human activities? Environmental impacts are not equivalent to mitigation policy impacts (the effects -- both positive and negative -- that arise from the implementation of laws, regulations, and initiatives aimed at managing and protecting the environment).
    \item \textbf{Options:} Yes, No
    \item \textbf{Examples:}
    \begin{itemize}
        \item Positive examples (code `Yes'):
        \begin{itemize}
            \item ``The threat that carbon emissions oppose to our way of society by inducing large scale climate change is a problem we must face up to very quickly.''
            \item ``But we still have major problems: in many cities, air quality is badly polluted and people suffer from fine particulate emissions.''
        \end{itemize}
        \item Negative examples (code `No'):
        \begin{itemize}
            \item ``The implementation of the Paris Agreement will also play an important role in terms of limiting the increase in global average temperature to below 2°C pre-industrial levels.''
            \item ``On the one hand, the European fleet needs access to new fishing grounds, particularly to be able to track the migration of certain fish species, notably tuna.''
        \end{itemize}
    \end{itemize}
\end{itemize}

\subsubsection{Specificity}
\begin{itemize}
    \item \textbf{Question:} Are the environmental impacts described as having specific settings (impacting specifically mentioned individuals, groups, industries, or regions) or universal settings (impacting humanity as a whole or other lifeforms)? If in doubt or no clear explicit or implicit setting, code `Universal'.
    \item \textbf{Options:} Specific, Universal
    \item \textbf{Examples:}
    \begin{itemize}
        \item Code `Specific':
        \begin{itemize}
            \item ``One hundred billion plastic bags are consumed in Europe every year, and their excessive use has a disastrous impact on the environment.''
            \item ``Just tonight, in Romania, one of the most beautiful waterfalls, the Bigăr waterfall, collapsed under human action.''
            \item ``In the first half of 2022 alone, 3,750 square kilometres of rainforest have been destroyed -- gone forever.''
        \end{itemize}
        \item Code `Universal':
        \begin{itemize}
            \item ``However, it is a group of substances that are suspected of impairing liver and thyroid function and causing cancer.''
            \item ``In the near future, this scenario will be common, with fires recurring, increasing in scale, intensity and frequency.''
        \end{itemize}
    \end{itemize}
\end{itemize}

\subsubsection{Proximity}
\begin{itemize}
    \item \textbf{Question:} Are the environmental impacts portrayed as proximate or distant? Proximate means impacting EU nationals, EU member states, or specific communities within the bloc, currently or in the near future. Distant means impacting non-EU states or communities, or future generations.
    \item \textbf{Options:} Proximate, Distant
    \item \textbf{Examples:}
    \begin{itemize}
        \item Code `Proximate':
        \begin{itemize}
            \item ``According to a Commission impact assessment, the health costs caused by air pollution across Europe amount to around EUR 940 million per year.''
            \item ``Indeed, 84\% of Europeans have expressed their fears regarding the impact of chemicals on their health.''
        \end{itemize}
        \item Code `Distant':
        \begin{itemize}
            \item ``In the first half of 2022 alone, 3,750 square kilometres of rainforest have been destroyed -- gone forever.''
            \item ``This illegal form of fishing has emerged, particularly since the civil war, and has an adverse impact not only from an environmental point of view, but also financially damages Liberian fishermen and the state budget.''
        \end{itemize}
    \end{itemize}
\end{itemize}
\newpage

\subsection{Economic Sentiment}
Read the newspaper article excerpt and assess what indication it gives about how the US economy is performing. Rate the sentiment as either negative or positive only. If the text contains typos or is corrupted, do your best to interpret the intended message.

\subsubsection{Positivity}
\begin{itemize}
    \item \textbf{Question:} Based on the indication the article gives about how the US economy is performing, rate whether the indication is negative or positive. Consider the overall tone and content regarding US economic performance mentioned in the excerpt.
    \item \textbf{Options:} Negative, Positive
    \item \textbf{Examples:}
    \begin{itemize}
        \item Code `Negative':
        \begin{itemize}
            \item ``STAGNANT WAGES POSE ADDED RISKS TO WEAK ECONOMY: CONSUMERS FEEL A PINCH Bigger Deductions for Medical Coverage and Less Overtime Put Dent in Paychecks Stagnant Wages Add to Risk for Economy. Although the recession has ended, the wages of more than 100 million workers are still stagnant, endangering the consumer spending that sustains the fragile recovery [\ldots]''
            \item ``Trade Deficit Widened by Oil Imports: Trade Deficit Widened by Oil Imports. WASHINGTON, March 20 --- The trade deficit widened to \$9.3 billion in January as oil imports surged to refill stocks depleted during December's record cold [\ldots]''
        \end{itemize}
        \item Code `Positive':
        \begin{itemize}
            \item ``Market Undergoes Mild Consolidation As Economy Rises: THE WEEK IN FINANCE. ALMOST all the background news last week was decidedly constructive, but the stock market failed to take heart from it and instead endured a mild period of consolidation [\ldots]''
            \item ``Investors Regain Optimism as Crude Oil Prices Decline: Key Rates. Stocks climbed yesterday as a drop in crude oil prices from their peak eased investors' worries that high fuel costs would crimp consumer spending and hurt company profits [\ldots]''
        \end{itemize}
    \end{itemize}
\end{itemize}
\newpage

\subsection{Manifesto Topic}
Classify each text into one of the domains below.

\subsubsection{Domain}
\begin{itemize}
    \item \textbf{Question:} Select the domain that best matches the content of the text.
    \item \textbf{Options:}
    \begin{itemize}
        \item \textbf{No Domain}
        \item \textbf{External Relations:} Foreign Special Relationships, Anti-Imperialism, Military, Peace, Internationalism, European Community/Union
        \item \textbf{Freedom And Democracy:} Freedom And Human Rights, Democracy, Constitutionalism
        \item \textbf{Political System:} Decentralization, Governmental And Administrative Efficiency, Political Corruption, Political Authority
        \item \textbf{Economy:} Free Market Economy, Incentives, Market Regulation, Economic Planning, Corporatism, Protectionism, Economic Goals, Keynesian Demand Management, Economic Growth, Technology And Infrastructure, Controlled Economy, Nationalization, Economic Orthodoxy, Marxist Analysis, Anti-Growth Economy
        \item \textbf{Welfare And Quality Of Life:} Environmental Protection, Culture, Equality, Welfare State Expansion, Welfare State Limitation, Education Expansion, Education Limitation
        \item \textbf{Fabric Of Society:} National Way Of Life, Traditional Morality, Law And Order, Civic Mindedness, Multiculturalism
        \item \textbf{Social Groups:} Labour Groups, Agriculture And Farmers, Middle Class And Professional Groups, Underprivileged Minority Groups, Non-economic Demographic Groups
    \end{itemize}
    \item \textbf{Examples:}
    \begin{itemize}
        \item Code `External Relations':
        \begin{itemize}
            \item ``The United States government's policy is that there is one China, as reflected in the three communiqu\'{e}s and the Taiwan Relations Act.''
            \item ``and Iraqis allowed their self-determination.''
            \item ``Roaming Charges within the EU: Fine Gael strongly supports the principle that roaming charges should be harmonised across the EU and we will work in Europe to speed up progress in this regard.''
        \end{itemize}
        \item Code `Freedom And Democracy':
        \begin{itemize}
            \item ``The rights of citizenship do not stop at the ballot box.''
            \item ``These goals require that we build a democratic developmental state capable of mobilising all sectors and boldly intervening in the economy in favour of workers and the poor.''
            \item ``You can vote to retain the old First Past the Post system, or you can vote for the Mixed Member Proportional system - MMP.''
        \end{itemize}
        \item Code `Political System':
        \begin{itemize}
            \item ``allow no major reorganisation of local government in our first term in office.''
            \item ``The restructured Regional Development Australia network will navigate federal government funding programs, according to our fair share commitment, including a restructured Better Regions Program.''
            \item ``Unlike Labour, I'm not prepared to mislead the New Zealand public about the situation this country finds itself in.''
        \end{itemize}
        \item Code `Economy':
        \begin{itemize}
            \item ``who maintain their homes well and have lower maintenance costs as a result.''
            \item ``Develop a national plan to address the impacts of climate change on rural communities and regional industries.''
            \item ``Moreover, the inflation tax is regressive.''
            \item ``The measures we propose as alternatives to austerity will, by halting and reversing the cuts to public services, restore lost jobs and create new ones.''
            \item ``Our plan can be entirely implemented within the ten-year transport budget set out in the draft 2015-2025 Government Policy Statement on Transport Funding.''
        \end{itemize}
        \item Code `Welfare And Quality Of Life':
        \begin{itemize}
            \item ``Meanwhile, to help the poorest students now, we will immediately restore maintenance grants.''
            \item ``At present New Zealanders are amongst the highest per capita emitters in the world.''
            \item ``This new funding will not impact those students hoping to study in the humanities.''
        \end{itemize}
        \item Code `Fabric Of Society':
        \begin{itemize}
            \item ``and give Welsh speakers the right to use their language at all levels and for its status to be internationally recognised.''
            \item ``Every percentage drop represents an assault on the ties that bind us together.''
            \item ``Since then we have seen our nation build on a diverse heritage of cultures that has contributed to our proud nation.''
            \item ``The referendum on marriage equality was a historic victory for the rights of gay and lesbian people.''
        \end{itemize}
        \item Code `Social Groups':
        \begin{itemize}
            \item ``It also means providing support that has been proven to work, like work experience placements that help them get a first foot on the career ladder.''
            \item ``We will reduce the tax burden for those on low and middle incomes Income tax and USC.''
            \item ``Women in Politics: Fine Gael recognises that there needs to be a substantial increase in the number of women in politics.''
        \end{itemize}
    \end{itemize}
\end{itemize}

\newpage

\section{Performance Metrics Tables}\label{app:performance-metrics}
\input{metrics_appendix}


\end{document}

%% file: metrics_appendix.tex

\subsection{Model Choice}

\begin{longtable}[t]{llrrrrrr}
\caption{\label{tab:appendix-model-choice-comprehensive-metrics}Task-level performance metrics for the model choice comparison, reporting F1, accuracy, precision, recall, Cohen's $\kappa$, and Krippendorff's $\alpha $ for each model.}\\
\toprule
Model & Task & F1 & Accuracy & Precision & Recall & $\kappa$ & $\alpha$\\
\midrule
DeepSeek-R1 & approval & 0.357 & 0.414 & 0.476 & 0.355 & 0.225 & 0.538\\
DeepSeek-R1 & psych-dist & 0.393 & 0.948 & 0.415 & 0.376 & 0.643 & 0.656\\
DeepSeek-R1 & sentiment & 0.735 & 0.743 & 0.732 & 0.741 & 0.471 & 0.470\\
DeepSeek-R1 & topic-8 & 0.234 & 0.496 & 0.282 & 0.248 & 0.378 & 0.385\\
GPT-OSS & approval & 0.427 & 0.477 & 0.481 & 0.418 & 0.285 & 0.571\\
GPT-OSS & psych-dist & 0.699 & 0.935 & 0.726 & 0.681 & 0.547 & 0.564\\
GPT-OSS & sentiment & 0.700 & 0.719 & 0.703 & 0.698 & 0.400 & 0.401\\
GPT-OSS & topic-8 & 0.359 & 0.496 & 0.416 & 0.391 & 0.385 & 0.447\\
Gemma 3 & approval & 0.378 & 0.413 & 0.511 & 0.377 & 0.234 & 0.562\\
Gemma 3 & psych-dist & 0.678 & 0.893 & 0.638 & 0.769 & 0.500 & 0.499\\
Gemma 3 & sentiment & 0.714 & 0.729 & 0.714 & 0.714 & 0.427 & 0.428\\
Gemma 3 & topic-8 & 0.169 & 0.448 & 0.225 & 0.177 & 0.327 & 0.322\\
LLaMA 3.1 & approval & 0.434 & 0.473 & 0.505 & 0.422 & 0.280 & 0.606\\
LLaMA 3.1 & psych-dist & 0.490 & 0.848 & 0.454 & 0.629 & 0.434 & 0.333\\
LLaMA 3.1 & sentiment & 0.488 & 0.740 & 0.487 & 0.491 & 0.466 & 0.467\\
LLaMA 3.1 & topic-8 & 0.146 & 0.431 & 0.188 & 0.157 & 0.321 & 0.344\\
Qwen 2.5 & approval & 0.380 & 0.403 & 0.499 & 0.382 & 0.209 & 0.576\\
Qwen 2.5 & psych-dist & 0.687 & 0.941 & 0.812 & 0.626 & 0.516 & 0.528\\
Qwen 2.5 & sentiment & 0.365 & 0.733 & 0.366 & 0.368 & 0.461 & 0.460\\
Qwen 2.5 & topic-8 & 0.175 & 0.413 & 0.223 & 0.174 & 0.313 & 0.392\\
Qwen 3 & approval & 0.419 & 0.461 & 0.475 & 0.408 & 0.268 & 0.560\\
Qwen 3 & psych-dist & 0.720 & 0.925 & 0.693 & 0.758 & 0.566 & 0.576\\
Qwen 3 & sentiment & 0.721 & 0.726 & 0.722 & 0.733 & 0.447 & 0.443\\
Qwen 3 & topic-8 & 0.304 & 0.507 & 0.347 & 0.322 & 0.394 & 0.439\\
\bottomrule
\end{longtable}

\subsection{Model Size}

\begin{longtable}[t]{llrrrrrr}
\caption{\label{tab:appendix-model-size-comprehensive-metrics}Task-level performance metrics for the model size comparison, reporting F1, accuracy, precision, recall, Cohen's $\kappa $, and Krippendorff's $\alpha $ for each model variant.}\\
\toprule
Model & Task & F1 & Accuracy & Precision & Recall & $\kappa$ & $\alpha$\\
\midrule
Deepseek-R1 1.5B & approval & 0.177 & 0.219 & 0.217 & 0.220 & 0.024 & -0.008\\
Deepseek-R1 1.5B & psych-dist & 0.048 & 0.779 & 0.050 & 0.047 & 0.066 & 0.031\\
Deepseek-R1 1.5B & sentiment & 0.015 & 0.486 & 0.018 & 0.013 & 0.183 & -0.056\\
Deepseek-R1 1.5B & topic-8 & 0.002 & 0.153 & 0.004 & 0.003 & 0.071 & 0.025\\
Deepseek-R1 8B & approval & 0.335 & 0.416 & 0.435 & 0.338 & 0.216 & 0.503\\
Deepseek-R1 8B & psych-dist & 0.305 & 0.908 & 0.322 & 0.301 & 0.437 & 0.411\\
Deepseek-R1 8B & sentiment & 0.679 & 0.700 & 0.682 & 0.677 & 0.358 & 0.358\\
Deepseek-R1 8B & topic-8 & 0.292 & 0.481 & 0.328 & 0.312 & 0.369 & 0.362\\
Deepseek-R1 14B & approval & 0.374 & 0.414 & 0.493 & 0.361 & 0.216 & 0.500\\
Deepseek-R1 14B & psych-dist & 0.237 & 0.927 & 0.259 & 0.228 & 0.518 & 0.522\\
Deepseek-R1 14B & sentiment & 0.069 & 0.686 & 0.070 & 0.068 & 0.377 & 0.048\\
Deepseek-R1 14B & topic-8 & 0.045 & 0.455 & 0.053 & 0.047 & 0.347 & 0.528\\
Deepseek-R1 32B & approval & 0.396 & 0.450 & 0.453 & 0.385 & 0.256 & 0.587\\
Deepseek-R1 32B & psych-dist & 0.290 & 0.929 & 0.299 & 0.283 & 0.558 & 0.525\\
Deepseek-R1 32B & sentiment & 0.179 & 0.724 & 0.180 & 0.179 & 0.434 & 0.254\\
Deepseek-R1 32B & topic-8 & 0.049 & 0.484 & 0.057 & 0.052 & 0.377 & 0.553\\
Deepseek-R1 70B & approval & 0.357 & 0.414 & 0.476 & 0.355 & 0.225 & 0.538\\
Deepseek-R1 70B & psych-dist & 0.393 & 0.948 & 0.415 & 0.376 & 0.643 & 0.656\\
Deepseek-R1 70B & sentiment & 0.735 & 0.743 & 0.732 & 0.741 & 0.471 & 0.470\\
Deepseek-R1 70B & topic-8 & 0.234 & 0.496 & 0.282 & 0.248 & 0.378 & 0.385\\
Gemma3 270M & approval & 0.152 & 0.259 & 0.164 & 0.203 & -0.008 & 0.023\\
Gemma3 270M & psych-dist & 0.063 & 0.528 & 0.064 & 0.064 & 0.001 & -0.152\\
Gemma3 270M & sentiment & 0.054 & 0.276 & 0.129 & 0.042 & 0.037 & 0.078\\
Gemma3 270M & topic-8 & 0.000 & 0.003 & 0.000 & 0.000 & 0.000 & -0.003\\
Gemma3 1B & approval & 0.118 & 0.169 & 0.158 & 0.214 & 0.005 & -0.265\\
Gemma3 1B & psych-dist & 0.213 & 0.615 & 0.205 & 0.222 & 0.000 & -0.097\\
Gemma3 1B & sentiment & 0.282 & 0.524 & 0.337 & 0.246 & 0.194 & 0.318\\
Gemma3 1B & topic-8 & 0.016 & 0.088 & 0.047 & 0.022 & 0.017 & -0.038\\
Gemma3 4B & approval & 0.297 & 0.369 & 0.364 & 0.308 & 0.162 & 0.380\\
Gemma3 4B & psych-dist & 0.568 & 0.844 & 0.535 & 0.709 & 0.381 & 0.395\\
Gemma3 4B & sentiment & 0.689 & 0.710 & 0.693 & 0.687 & 0.378 & 0.379\\
Gemma3 4B & topic-8 & 0.065 & 0.351 & 0.110 & 0.082 & 0.182 & 0.287\\
Gemma3 12B & approval & 0.386 & 0.453 & 0.486 & 0.375 & 0.263 & 0.588\\
Gemma3 12B & psych-dist & 0.655 & 0.886 & 0.617 & 0.741 & 0.466 & 0.466\\
Gemma3 12B & sentiment & 0.709 & 0.736 & 0.724 & 0.703 & 0.422 & 0.419\\
Gemma3 12B & topic-8 & 0.259 & 0.455 & 0.351 & 0.291 & 0.337 & 0.345\\
Gemma3 27B & approval & 0.378 & 0.413 & 0.511 & 0.377 & 0.234 & 0.562\\
Gemma3 27B & psych-dist & 0.678 & 0.893 & 0.638 & 0.769 & 0.500 & 0.499\\
Gemma3 27B & sentiment & 0.714 & 0.729 & 0.714 & 0.714 & 0.427 & 0.428\\
Gemma3 27B & topic-8 & 0.169 & 0.448 & 0.225 & 0.177 & 0.327 & 0.322\\
Qwen3 600M & approval & 0.133 & 0.215 & 0.192 & 0.229 & 0.039 & -0.021\\
Qwen3 600M & psych-dist & 0.125 & 0.707 & 0.131 & 0.153 & 0.156 & 0.098\\
Qwen3 600M & sentiment & 0.413 & 0.619 & 0.453 & 0.444 & 0.294 & 0.239\\
Qwen3 600M & topic-8 & 0.058 & 0.251 & 0.092 & 0.080 & 0.139 & 0.117\\
Qwen3 1.7B & approval & 0.215 & 0.293 & 0.269 & 0.279 & 0.093 & 0.172\\
Qwen3 1.7B & psych-dist & 0.594 & 0.902 & 0.582 & 0.621 & 0.474 & 0.459\\
Qwen3 1.7B & sentiment & 0.688 & 0.690 & 0.696 & 0.707 & 0.388 & 0.377\\
Qwen3 1.7B & topic-8 & 0.169 & 0.308 & 0.230 & 0.174 & 0.198 & 0.134\\
Qwen3 4B & approval & 0.382 & 0.438 & 0.418 & 0.377 & 0.233 & 0.406\\
Qwen3 4B & psych-dist & 0.652 & 0.936 & 0.701 & 0.620 & 0.530 & 0.522\\
Qwen3 4B & sentiment & 0.292 & 0.738 & 0.293 & 0.290 & 0.454 & 0.437\\
Qwen3 4B & topic-8 & 0.147 & 0.505 & 0.166 & 0.158 & 0.396 & 0.411\\
Qwen3 8B & approval & 0.349 & 0.440 & 0.462 & 0.346 & 0.234 & 0.533\\
Qwen3 8B & psych-dist & 0.599 & 0.930 & 0.654 & 0.568 & 0.493 & 0.492\\
Qwen3 8B & sentiment & 0.701 & 0.710 & 0.699 & 0.707 & 0.403 & 0.402\\
Qwen3 8B & topic-8 & 0.301 & 0.450 & 0.361 & 0.325 & 0.341 & 0.368\\
Qwen3 14B & approval & 0.435 & 0.466 & 0.521 & 0.414 & 0.275 & 0.560\\
Qwen3 14B & psych-dist & 0.423 & 0.937 & 0.470 & 0.397 & 0.523 & 0.480\\
Qwen3 14B & sentiment & 0.717 & 0.733 & 0.718 & 0.716 & 0.435 & 0.435\\
Qwen3 14B & topic-8 & 0.299 & 0.469 & 0.358 & 0.302 & 0.363 & 0.446\\
Qwen3 32B & approval & 0.419 & 0.461 & 0.475 & 0.408 & 0.268 & 0.560\\
Qwen3 32B & psych-dist & 0.720 & 0.925 & 0.693 & 0.758 & 0.566 & 0.576\\
Qwen3 32B & sentiment & 0.721 & 0.726 & 0.722 & 0.733 & 0.447 & 0.443\\
Qwen3 32B & topic-8 & 0.304 & 0.507 & 0.347 & 0.322 & 0.394 & 0.439\\
\bottomrule
\end{longtable}

\subsection{Zero-Shot vs. Few-Shot}

\begin{longtable}[t]{lllrrrrrr}
\caption{\label{tab:appendix-fewshot-comprehensive-metrics}Task-level performance metrics for the zero-shot and few-shot comparison, reporting F1, accuracy, precision, recall, Cohen's $\kappa $, and Krippendorff's $\alpha $ by model and learning approach.}\\
\toprule
Model & \shortstack{Learning\\Approach} & Task & F1 & Accuracy & Precision & Recall & $\kappa$ & $\alpha$\\
\midrule
DeepSeek-R1 & Few-Shot & approval & 0.379 & 0.391 & 0.477 & 0.375 & 0.216 & 0.532\\
DeepSeek-R1 & Few-Shot & psych-dist & 0.677 & 0.937 & 0.646 & 0.719 & 0.640 & 0.649\\
DeepSeek-R1 & Few-Shot & sentiment & 0.713 & 0.729 & 0.713 & 0.712 & 0.426 & 0.427\\
DeepSeek-R1 & Few-Shot & topic-8 & 0.163 & 0.465 & 0.212 & 0.181 & 0.334 & 0.366\\
DeepSeek-R1 & Zero-Shot & approval & 0.357 & 0.414 & 0.476 & 0.355 & 0.225 & 0.538\\
DeepSeek-R1 & Zero-Shot & psych-dist & 0.393 & 0.948 & 0.415 & 0.376 & 0.643 & 0.656\\
DeepSeek-R1 & Zero-Shot & sentiment & 0.735 & 0.743 & 0.732 & 0.741 & 0.471 & 0.470\\
DeepSeek-R1 & Zero-Shot & topic-8 & 0.234 & 0.496 & 0.282 & 0.248 & 0.378 & 0.385\\
GPT-OSS & Few-Shot & approval & 0.424 & 0.453 & 0.453 & 0.417 & 0.267 & 0.582\\
GPT-OSS & Few-Shot & psych-dist & 0.551 & 0.914 & 0.522 & 0.595 & 0.532 & 0.474\\
GPT-OSS & Few-Shot & sentiment & 0.712 & 0.729 & 0.713 & 0.710 & 0.423 & 0.424\\
GPT-OSS & Few-Shot & topic-8 & 0.357 & 0.482 & 0.441 & 0.387 & 0.367 & 0.416\\
GPT-OSS & Zero-Shot & approval & 0.427 & 0.477 & 0.481 & 0.418 & 0.285 & 0.571\\
GPT-OSS & Zero-Shot & psych-dist & 0.699 & 0.935 & 0.726 & 0.681 & 0.547 & 0.564\\
GPT-OSS & Zero-Shot & sentiment & 0.700 & 0.719 & 0.703 & 0.698 & 0.400 & 0.401\\
GPT-OSS & Zero-Shot & topic-8 & 0.359 & 0.496 & 0.416 & 0.391 & 0.385 & 0.447\\
Gemma 3 & Few-Shot & approval & 0.422 & 0.412 & 0.504 & 0.420 & 0.244 & 0.571\\
Gemma 3 & Few-Shot & psych-dist & 0.617 & 0.835 & 0.574 & 0.807 & 0.394 & 0.395\\
Gemma 3 & Few-Shot & sentiment & 0.703 & 0.719 & 0.703 & 0.702 & 0.406 & 0.406\\
Gemma 3 & Few-Shot & topic-8 & 0.178 & 0.378 & 0.301 & 0.207 & 0.253 & 0.288\\
Gemma 3 & Zero-Shot & approval & 0.378 & 0.413 & 0.511 & 0.377 & 0.234 & 0.562\\
Gemma 3 & Zero-Shot & psych-dist & 0.678 & 0.893 & 0.638 & 0.769 & 0.500 & 0.499\\
Gemma 3 & Zero-Shot & sentiment & 0.714 & 0.729 & 0.714 & 0.714 & 0.427 & 0.428\\
Gemma 3 & Zero-Shot & topic-8 & 0.169 & 0.448 & 0.225 & 0.177 & 0.327 & 0.322\\
LLaMA 3.1 & Few-Shot & approval & 0.461 & 0.484 & 0.498 & 0.459 & 0.309 & 0.652\\
LLaMA 3.1 & Few-Shot & psych-dist & 0.572 & 0.787 & 0.543 & 0.809 & 0.335 & 0.327\\
LLaMA 3.1 & Few-Shot & sentiment & 0.474 & 0.717 & 0.475 & 0.475 & 0.421 & 0.427\\
LLaMA 3.1 & Few-Shot & topic-8 & 0.098 & 0.415 & 0.124 & 0.111 & 0.313 & 0.335\\
LLaMA 3.1 & Zero-Shot & approval & 0.434 & 0.473 & 0.505 & 0.422 & 0.280 & 0.606\\
LLaMA 3.1 & Zero-Shot & psych-dist & 0.490 & 0.848 & 0.454 & 0.629 & 0.434 & 0.333\\
LLaMA 3.1 & Zero-Shot & sentiment & 0.488 & 0.740 & 0.487 & 0.491 & 0.466 & 0.467\\
LLaMA 3.1 & Zero-Shot & topic-8 & 0.146 & 0.431 & 0.188 & 0.157 & 0.321 & 0.344\\
Qwen 2.5 & Few-Shot & approval & 0.424 & 0.400 & 0.548 & 0.425 & 0.222 & 0.592\\
Qwen 2.5 & Few-Shot & psych-dist & 0.736 & 0.926 & 0.699 & 0.788 & 0.588 & 0.596\\
Qwen 2.5 & Few-Shot & sentiment & 0.286 & 0.710 & 0.291 & 0.281 & 0.422 & 0.417\\
Qwen 2.5 & Few-Shot & topic-8 & 0.200 & 0.386 & 0.276 & 0.205 & 0.290 & 0.425\\
Qwen 2.5 & Zero-Shot & approval & 0.380 & 0.403 & 0.499 & 0.382 & 0.209 & 0.576\\
Qwen 2.5 & Zero-Shot & psych-dist & 0.687 & 0.941 & 0.812 & 0.626 & 0.516 & 0.528\\
Qwen 2.5 & Zero-Shot & sentiment & 0.365 & 0.733 & 0.366 & 0.368 & 0.461 & 0.460\\
Qwen 2.5 & Zero-Shot & topic-8 & 0.175 & 0.413 & 0.223 & 0.174 & 0.313 & 0.392\\
Qwen 3 & Few-Shot & approval & 0.439 & 0.456 & 0.456 & 0.433 & 0.277 & 0.612\\
Qwen 3 & Few-Shot & psych-dist & 0.732 & 0.935 & 0.750 & 0.717 & 0.576 & 0.577\\
Qwen 3 & Few-Shot & sentiment & 0.723 & 0.736 & 0.722 & 0.725 & 0.447 & 0.447\\
Qwen 3 & Few-Shot & topic-8 & 0.160 & 0.488 & 0.191 & 0.173 & 0.370 & 0.394\\
Qwen 3 & Zero-Shot & approval & 0.419 & 0.461 & 0.475 & 0.408 & 0.268 & 0.560\\
Qwen 3 & Zero-Shot & psych-dist & 0.720 & 0.925 & 0.693 & 0.758 & 0.566 & 0.576\\
Qwen 3 & Zero-Shot & sentiment & 0.721 & 0.726 & 0.722 & 0.733 & 0.447 & 0.443\\
Qwen 3 & Zero-Shot & topic-8 & 0.304 & 0.507 & 0.347 & 0.322 & 0.394 & 0.439\\
\bottomrule
\end{longtable}

\subsection{Prompt Style}

\begin{longtable}[t]{lllrrrrrr}
\caption{\label{tab:appendix-prompt-style-comprehensive-metrics}Task-level performance metrics for the prompt style comparison, reporting F1, accuracy, precision, recall, Cohen's $\kappa $, and Krippendorff's $\alpha $ by model and prompt style.}\\
\toprule
Model & \shortstack{Prompt\\Style} & Task & F1 & Accuracy & Precision & Recall & $\kappa$ & $\alpha$\\
\midrule
DeepSeek-R1 & CoT & approval & 0.429 & 0.433 & 0.520 & 0.418 & 0.261 & 0.563\\
DeepSeek-R1 & CoT & psych-dist & 0.626 & 0.905 & 0.594 & 0.678 & 0.516 & 0.499\\
DeepSeek-R1 & CoT & sentiment & 0.474 & 0.719 & 0.479 & 0.469 & 0.418 & 0.401\\
DeepSeek-R1 & CoT & topic-8 & 0.166 & 0.468 & 0.215 & 0.181 & 0.337 & 0.465\\
DeepSeek-R1 & persona & approval & 0.389 & 0.394 & 0.462 & 0.384 & 0.215 & 0.525\\
DeepSeek-R1 & persona & psych-dist & 0.744 & 0.935 & 0.716 & 0.783 & 0.609 & 0.634\\
DeepSeek-R1 & persona & sentiment & 0.717 & 0.731 & 0.716 & 0.718 & 0.434 & 0.435\\
DeepSeek-R1 & persona & topic-8 & 0.149 & 0.459 & 0.201 & 0.162 & 0.327 & 0.460\\
DeepSeek-R1 & standard & approval & 0.379 & 0.391 & 0.477 & 0.375 & 0.216 & 0.532\\
DeepSeek-R1 & standard & psych-dist & 0.677 & 0.937 & 0.646 & 0.719 & 0.640 & 0.649\\
DeepSeek-R1 & standard & sentiment & 0.713 & 0.729 & 0.713 & 0.712 & 0.426 & 0.427\\
DeepSeek-R1 & standard & topic-8 & 0.163 & 0.465 & 0.212 & 0.181 & 0.334 & 0.366\\
GPT-OSS & CoT & approval & 0.447 & 0.456 & 0.485 & 0.430 & 0.269 & 0.597\\
GPT-OSS & CoT & psych-dist & 0.616 & 0.923 & 0.594 & 0.649 & 0.548 & 0.559\\
GPT-OSS & CoT & sentiment & 0.718 & 0.736 & 0.721 & 0.716 & 0.436 & 0.437\\
GPT-OSS & CoT & topic-8 & 0.363 & 0.497 & 0.429 & 0.391 & 0.381 & 0.419\\
GPT-OSS & persona & approval & 0.448 & 0.468 & 0.464 & 0.447 & 0.286 & 0.597\\
GPT-OSS & persona & psych-dist & 0.637 & 0.932 & 0.622 & 0.666 & 0.583 & 0.591\\
GPT-OSS & persona & sentiment & 0.689 & 0.710 & 0.693 & 0.687 & 0.378 & 0.379\\
GPT-OSS & persona & topic-8 & 0.357 & 0.478 & 0.439 & 0.385 & 0.361 & 0.409\\
GPT-OSS & standard & approval & 0.424 & 0.453 & 0.453 & 0.417 & 0.267 & 0.582\\
GPT-OSS & standard & psych-dist & 0.551 & 0.914 & 0.522 & 0.595 & 0.532 & 0.474\\
GPT-OSS & standard & sentiment & 0.712 & 0.729 & 0.713 & 0.710 & 0.423 & 0.424\\
GPT-OSS & standard & topic-8 & 0.357 & 0.482 & 0.441 & 0.387 & 0.367 & 0.416\\
Gemma 3 & CoT & approval & 0.424 & 0.429 & 0.447 & 0.426 & 0.257 & 0.578\\
Gemma 3 & CoT & psych-dist & 0.567 & 0.786 & 0.539 & 0.773 & 0.316 & 0.301\\
Gemma 3 & CoT & sentiment & 0.692 & 0.712 & 0.695 & 0.690 & 0.384 & 0.385\\
Gemma 3 & CoT & topic-8 & 0.214 & 0.422 & 0.309 & 0.239 & 0.298 & 0.359\\
Gemma 3 & persona & approval & 0.418 & 0.413 & 0.496 & 0.418 & 0.244 & 0.573\\
Gemma 3 & persona & psych-dist & 0.633 & 0.846 & 0.583 & 0.816 & 0.417 & 0.411\\
Gemma 3 & persona & sentiment & 0.710 & 0.724 & 0.709 & 0.711 & 0.420 & 0.420\\
Gemma 3 & persona & topic-8 & 0.206 & 0.390 & 0.342 & 0.237 & 0.251 & 0.256\\
Gemma 3 & standard & approval & 0.422 & 0.412 & 0.504 & 0.420 & 0.244 & 0.571\\
Gemma 3 & standard & psych-dist & 0.617 & 0.835 & 0.574 & 0.807 & 0.394 & 0.395\\
Gemma 3 & standard & sentiment & 0.703 & 0.719 & 0.703 & 0.702 & 0.406 & 0.406\\
Gemma 3 & standard & topic-8 & 0.178 & 0.378 & 0.301 & 0.207 & 0.253 & 0.288\\
LLaMA 3.1 & CoT & approval & 0.395 & 0.463 & 0.450 & 0.382 & 0.275 & 0.599\\
LLaMA 3.1 & CoT & psych-dist & 0.644 & 0.863 & 0.594 & 0.797 & 0.437 & 0.446\\
LLaMA 3.1 & CoT & sentiment & 0.726 & 0.733 & 0.724 & 0.735 & 0.455 & 0.453\\
LLaMA 3.1 & CoT & topic-8 & 0.247 & 0.486 & 0.302 & 0.265 & 0.367 & 0.403\\
LLaMA 3.1 & persona & approval & 0.471 & 0.497 & 0.504 & 0.470 & 0.325 & 0.633\\
LLaMA 3.1 & persona & psych-dist & 0.231 & 0.706 & 0.236 & 0.336 & 0.246 & 0.185\\
LLaMA 3.1 & persona & sentiment & 0.363 & 0.731 & 0.364 & 0.362 & 0.448 & 0.461\\
LLaMA 3.1 & persona & topic-8 & 0.121 & 0.408 & 0.164 & 0.140 & 0.297 & 0.364\\
LLaMA 3.1 & standard & approval & 0.461 & 0.484 & 0.498 & 0.459 & 0.309 & 0.652\\
LLaMA 3.1 & standard & psych-dist & 0.572 & 0.787 & 0.543 & 0.809 & 0.335 & 0.327\\
LLaMA 3.1 & standard & sentiment & 0.474 & 0.717 & 0.475 & 0.475 & 0.421 & 0.427\\
LLaMA 3.1 & standard & topic-8 & 0.098 & 0.415 & 0.124 & 0.111 & 0.313 & 0.335\\
Qwen 2.5 & CoT & approval & 0.436 & 0.417 & 0.521 & 0.437 & 0.241 & 0.598\\
Qwen 2.5 & CoT & psych-dist & 0.749 & 0.932 & 0.729 & 0.774 & 0.606 & 0.610\\
Qwen 2.5 & CoT & sentiment & 0.365 & 0.724 & 0.370 & 0.361 & 0.452 & 0.477\\
Qwen 2.5 & CoT & topic-8 & 0.204 & 0.444 & 0.260 & 0.212 & 0.328 & 0.368\\
Qwen 2.5 & persona & approval & 0.429 & 0.415 & 0.523 & 0.430 & 0.240 & 0.619\\
Qwen 2.5 & persona & psych-dist & 0.741 & 0.928 & 0.705 & 0.790 & 0.598 & 0.604\\
Qwen 2.5 & persona & sentiment & 0.364 & 0.719 & 0.374 & 0.355 & 0.446 & 0.488\\
Qwen 2.5 & persona & topic-8 & 0.178 & 0.414 & 0.243 & 0.182 & 0.305 & 0.390\\
Qwen 2.5 & standard & approval & 0.424 & 0.400 & 0.548 & 0.425 & 0.222 & 0.592\\
Qwen 2.5 & standard & psych-dist & 0.736 & 0.926 & 0.699 & 0.788 & 0.588 & 0.596\\
Qwen 2.5 & standard & sentiment & 0.286 & 0.710 & 0.291 & 0.281 & 0.422 & 0.417\\
Qwen 2.5 & standard & topic-8 & 0.200 & 0.386 & 0.276 & 0.205 & 0.290 & 0.425\\
Qwen 3 & CoT & approval & 0.480 & 0.496 & 0.503 & 0.465 & 0.322 & 0.638\\
Qwen 3 & CoT & psych-dist & 0.735 & 0.929 & 0.718 & 0.756 & 0.576 & 0.573\\
Qwen 3 & CoT & sentiment & 0.741 & 0.752 & 0.739 & 0.743 & 0.482 & 0.483\\
Qwen 3 & CoT & topic-8 & 0.152 & 0.484 & 0.179 & 0.163 & 0.361 & 0.446\\
Qwen 3 & persona & approval & 0.461 & 0.476 & 0.470 & 0.457 & 0.302 & 0.637\\
Qwen 3 & persona & psych-dist & 0.724 & 0.929 & 0.716 & 0.734 & 0.558 & 0.553\\
Qwen 3 & persona & sentiment & 0.746 & 0.755 & 0.743 & 0.751 & 0.492 & 0.492\\
Qwen 3 & persona & topic-8 & 0.162 & 0.484 & 0.200 & 0.172 & 0.363 & 0.458\\
Qwen 3 & standard & approval & 0.439 & 0.456 & 0.456 & 0.433 & 0.277 & 0.612\\
Qwen 3 & standard & psych-dist & 0.732 & 0.935 & 0.750 & 0.717 & 0.576 & 0.577\\
Qwen 3 & standard & sentiment & 0.723 & 0.736 & 0.722 & 0.725 & 0.447 & 0.447\\
Qwen 3 & standard & topic-8 & 0.160 & 0.488 & 0.191 & 0.173 & 0.370 & 0.394\\
\bottomrule
\end{longtable}